\documentclass[runningheads]{llncs}
\pdfoutput=1
 
\usepackage{eccv}

\usepackage{colortbl}
\usepackage{adjustbox}
\usepackage{graphicx, amsmath, amssymb, caption, subcaption, multirow, overpic, textpos}
\newlength\savewidth\newcommand\shline{\noalign{\global\savewidth\arrayrulewidth
\global\arrayrulewidth 1pt}\hline\noalign{\global\arrayrulewidth\savewidth}}
\newcommand{\tablestyle}[2]{\setlength{\tabcolsep}{#1}\renewcommand{\arraystretch}{#2}\centering\footnotesize}
\renewcommand{\paragraph}[1]{\vspace{1.25mm}\noindent\textbf{#1}}

\newcolumntype{x}[1]{>{\centering\arraybackslash}p{#1pt}}
\newcolumntype{y}[1]{>{\raggedright\arraybackslash}p{#1pt}}
\newcolumntype{z}[1]{>{\raggedleft\arraybackslash}p{#1pt}}

\newcommand{\app}{\raise.17ex\hbox{$\scriptstyle\sim$}}

\definecolor{deemph}{gray}{0.6}

\definecolor{baselinecolor}{gray}{.9}
\newcommand{\baseline}[1]{\cellcolor{baselinecolor}{#1}}

\usepackage{marvosym}

\usepackage{eccvabbrv}
\usepackage{wrapfig,lipsum,booktabs}
\usepackage{graphicx}
\usepackage{booktabs}

\usepackage[accsupp]{axessibility}  

\usepackage{hyperref}

\usepackage{orcidlink}

\begin{document}

\title{Parameter-Efficient and Memory-Efficient Tuning for Vision Transformer:
A Disentangled Approach} 


\author{Taolin Zhang\inst{1}* \and
Jiawang Bai\inst{2}* \and
Zhihe Lu\inst{3} \and
Dongze Lian\inst{3} \and
Genping Wang$^{4,{\textrm{\Letter}}}$ \and
Xinchao Wang\inst{3,{\textrm{\Letter}}} \and
Shu-Tao Xia$^{1,5}$}

\authorrunning{Taolin Zhang, Jiawang Bai et al.}

\institute{$^1$Tsinghua Shenzhen International Graduate School, Tsinghua University \\
$^2$Tencent \quad $^3$National University of Singapore \\
$^4$Shenzhen Qiji Technology Co., Ltd. \\
$^5$Research Center of Artificial Intelligence, Peng Cheng Laboratory
{\tt ztl23@mails.tsinghua.edu.cn; jiawangbai@tencent.com;\\ zhihelu.academic@gmail.com;
dzlianx@gmail.com;\\ genpingwang@163.com; xinchao@nus.edu.sg; xiast@sz.tsinghua.edu.cn}
}
\renewcommand{\thefootnote}{\fnsymbol{footnote}}
\footnotetext{* Equal contribution.}
\footnotetext{\textrm{\Letter} Corresponding authors.}
\titlerunning{Parameter-Efficient and Memory-Efficient Tuning for Vision Transformer}

\maketitle

\begin{abstract}
Recent works on parameter-efficient transfer learning (PETL) show the potential to adapt a pre-trained Vision Transformer to downstream recognition tasks with only a few learnable parameters. 
However, since they usually insert new structures into the pre-trained model, entire intermediate features of that model are changed and thus need to be stored to be involved in back-propagation, resulting in memory-heavy training.
We solve this problem from a novel disentangled perspective, i.e., dividing PETL into two aspects: task-specific learning and pre-trained knowledge utilization.
Specifically, we synthesize the task-specific query with a learnable and lightweight module, which is independent of the pre-trained model. 
The synthesized query equipped with task-specific knowledge serves to extract the useful features for downstream tasks from the intermediate representations of the pre-trained model in a query-only manner.
Built upon these features, a customized classification head is proposed to make the prediction for the input sample.
Given that our method employs an extremely 
lightweight architecture and avoids the use of heavy intermediate features for running gradient descent, it demonstrates limited memory usage in training. 
Extensive experiments manifest that our method achieves state-of-the-art performance under memory constraints, showcasing its applicability in real-world situations.
The code is available at: \url{https://synqt.github.io/}.
\end{abstract}
\section{Introduction}

Motivated by the success of data-driven solutions, there has been an explosive growth in the size of vision models \cite{dosovitskiy2020image,liu2021swin, bai2022improving, wang2022image,kirillov2023segment, zhu2024vision, guo2024mambair} and the scale of accessible datasets like ImageNet \cite{deng2009imagenet}.
To unleash the power of large models, it becomes a tendency to first pre-train on a general and large-scale dataset, and then adapt them to downstream tasks \cite{kornblith2019better,shao2014transfer, wang2018deep}. 
The most obvious adaption strategy is to fully fine-tune all parameters of the pre-trained model. However, it necessitates that the device saves a dedicated set of model parameters for each single task, leading to substantial storage overhead \cite{zhai2022scaling}.

To improve the storage efficiency, parameter-efficient transfer learning (PETL) methods \cite{hu2021lora, chen2022adaptformer, jie2023fact, zhang2022NOAH, jie2023revisiting, jia2022visual} have been explored by tuning only
a small set of the parameters in adapting the large pre-trained models to downstream tasks. Typically, these methods insert small structures into the frozen pre-trained model and specify the newly added parameters as the only learnable ones. For instance, Visual Prompt Tuning (VPT) \cite{jia2022visual} prepends learnable tokens to the patch embeddings; AdaptFormer \cite{chen2022adaptformer} inserts a bottleneck-structured fully connected
layers parallel to the feed-forward network (FFN) of the original ViT. 
These studies demonstrate that PETL is a promising way to save storage resources. 

However, even though updating a small set of parameters, existing PETL methods still require a very high consumption of memory in training. 
The direct reason is that they need a full back-propagation throughout the entire backbone.
Given the fact that the model size is exponentially increasing, it becomes infeasible to apply existing PETL methods to the memory-constraint situations. Therefore, it is desirable to investigate memory-efficient PETL while maintaining performance on downstream tasks. 

In this work, we attribute the high memory cost of existing PETL methods in training to their intrinsic regime, \ie, entangling task-specific learning and pre-trained knowledge utilization together. It is usually achieved by integrating trainable structures into the pre-trained ViT and changing all intermediate features. Consequently, a great number of intermediate features need to be stored in training to run gradient descent. The aforementioned analysis motivates us to design a memory-efficient PETL method from a disentangled perspective.

\begin{figure*}[t]
    \centering
    \includegraphics[width=0.95\textwidth]{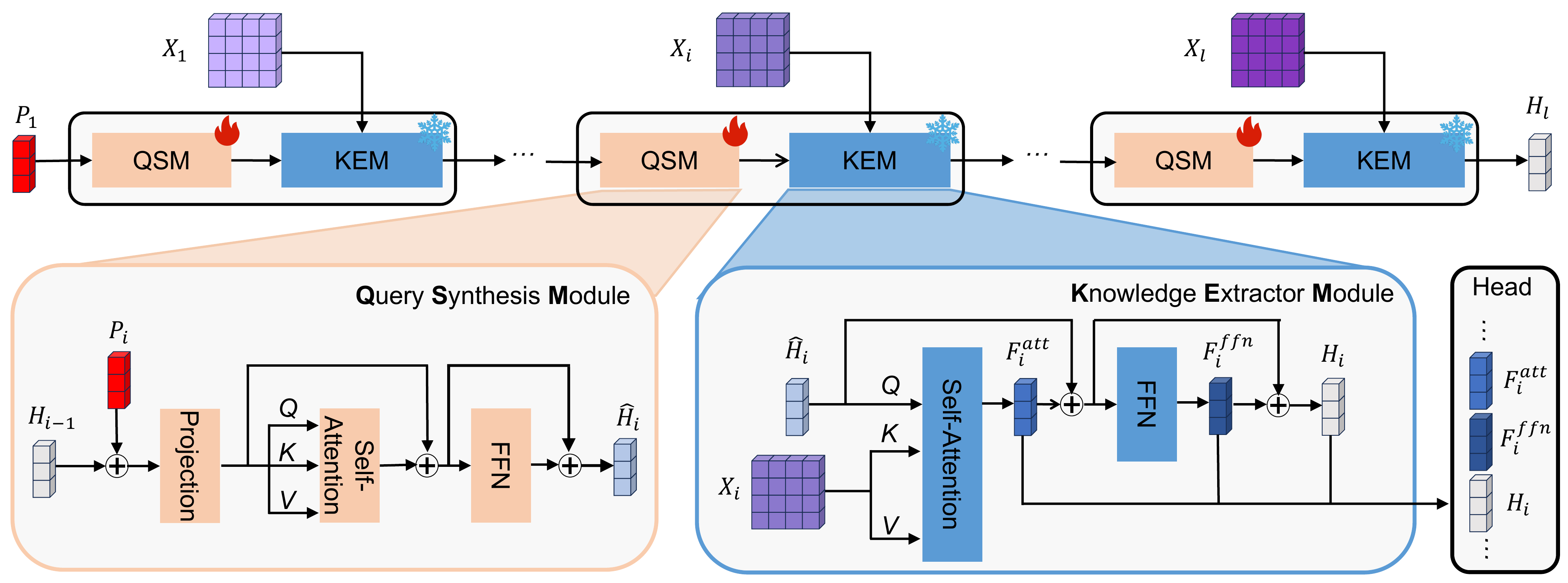}
    \caption{\textbf{Overall architecture of SynQT.} Each block in SynQT contains two key designs, including Query Synthesis Module (QSM) and Knowledge Extractor Module (KEM). QSM is a lightweight learnable module to capture task-specific knowledge and KEM reuses frozen weights in the pre-trained model for feature extraction. The rich and hierarchical features from KEMs are fed into the classification head.  }
    \label{fig:framework}
\end{figure*}

Accordingly, we propose \textbf{Syn}thesized \textbf{Q}uery \textbf{T}uning (SynQT), a memory-efficient PETL method for adapting pre-trained ViTs to downstream recognition tasks. 
Overall, as shown in Figure \ref{fig:framework}, SynQT takes the intermediate representations of the original pre-trained model as input and combines them with task-specific knowledge using stacked blocks.
We propose the \textbf{Q}uery \textbf{S}ynthesis \textbf{M}odule (QSM) and \textbf{K}nowledge \textbf{E}xtraction \textbf{M}odule (KEM) to accomplish the disentanglement.
To capture task-specific knowledge, SynQT employs a lightweight QSM, which is independent of the pre-trained model.
In QSM, an attention layer and an FFN are trained on the target downstream task to synthesize task-specific query. 
Then, to utilize the pre-trained knowledge, the synthesized query interacts with the intermediate representations of the original pre-trained model in a KEM. It outputs useful features for downstream recognition, where all parameters  are inherited from the pre-trained model and frozen. 
Relying on rich hierarchical features from KEMs, we propose a classification head, which adaptively aggregates these features conditioned on the input sample for prediction. 

The success of SynQT in saving memory in training is derived from two technical aspects. Firstly, we keep intermediate representations of the original pre-trained model intact and directly exploit them in a query-only manner, avoiding the use of heavy intermediate features for running gradient descent. Besides, our design including QSM and KEM is extremely lightweight. For instance,
we adopt bottleneck structures for Q-K-V projection \cite{vaswani2017attention} and a small number of tokens (lower than 4 in most cases) in QSM.

We validate SynQT on VTAB-1K \cite{zhai2019visual} benchmark, including 19 tasks from diverse domains. We also evaluate our method in few-shot learning using 5 fine-grained visual classification datasets. Extensive experiments show that SynQT can significantly reduce the memory footprint in training with promising accuracy. Compared to VPT-Deep, a powerful version of VPT that inserts prompts in each layer of ViT, SynQT saves 56\% memory consumption and achieves higher Top-1 accuracy on average. 
The contributions of our work are as follows:
 \begin{enumerate}
    \item We provide an insightful analysis of the high memory cost of existing PETL methods, $i.e.$, they entangle task-specific learning and pre-trained knowledge utilization together.   
    \item Inspired by the disentangled idea, we propose SynQT, which synthesizes task-specific query and then utilizes the pre-trained knowledge in a query-only manner.
    \item Extensive experiments show that SynQT is highly memory-efficient and effective, notably outperforming existing baselines under memory constraints. 
\end{enumerate}
\section{Related Work}
\noindent\textbf{Parameter-Efficient Transfer Learning (PETL).}
The idea of PETL initially shows great success in the field of natural language processing (NLP) \cite{brown2020language,hu2021lora,liu2022few,rebuffi2017learning}. 
It fine-tunes a small number of trainable parameters to transfer large pre-trained models to downstream tasks, significantly reducing storage costs by avoiding a copy of model parameters for each new task.
Due to the increased size of pre-trained vision models, especially for ViT \cite{dosovitskiy2020image}, PETL has been introduced to computer vision and attracted much attention \cite{tu2023visual,chen2022adaptformer, lian2022scaling,jia2022visual,zhou2022learning,zhou2022conditional,zang2022unified,pfeiffer2020adapterfusion, sung2022vl, li2024graphadapter, zha2023instance, guo2023adaptir,bai2024badclip,yang2024not,lu2024beyond}. 
To reduce the number of fine-tuned parameters, BitFit \cite{zaken2021bitfit} only fine-tunes the bias terms and freezes most of the
network, resulting in comparable performance to full tuning. 
VPT \cite{jia2022visual} introduces learnable tokens (i.e., prompts \cite{liu2023pre, li2021prefix, lester2021power, liu2021p}) to each layer of ViT.
SSF \cite{lian2022scaling} fine-tunes scale and shift factors in each layer to adapt original features to the target domain.  
Some other works \cite{houlsby2019parameter,hu2021lora,chen2022adaptformer} propose to insert additional trainable structures into attention blocks of the pre-trained Transformer.
For instance, LoRA \cite{hu2021lora} observes that the learned over-parametrized models reside on a low-rank dimension and further injects trainable rank decomposition parameters into each layer of the Transformer; AdaptFormer \cite{chen2022adaptformer} adds parallel lightweight module to FFN.
Furthermore, for extreme parameter efficiency, a tensorization-decomposition framework \cite{jie2023fact} is exploited to store the weight increments, and the low-bit adapter \cite{jie2023revisiting} is proposed to reduce the precision redundancy. 
However, how to improve memory efficiency in training is still an under-explored problem in PETL, as discussed below.

\noindent\textbf{Memory-Efficient PETL.}
Although existing PETL methods have shown promising performance in transfer learning while saving most trainable parameters, they still suffer from issues of high memory consumption in training. The full back-propagation throughout the entire backbone prevents traditional PETL methods from being applied in memory-constraint situations. 
Recently, some recent works \cite{sung2022lst,tu2023visual,evci2022head2toe} have made several attempts to address this issue, whose key idea is to leverage the power of intermediate representations of the pre-trained model and thus avoid full back-propagation. 
In NLP, LST \cite{sung2022lst} proposes a trainable lightweight structure in the sideway for memory-saving during training. 
To adapt vision models, Head2Toe \cite{evci2022head2toe} selects features from all layers of the source model to feed into a classification head. VQT \cite{tu2023visual} inserts trainable tokens in the query space to capture useful information from the pre-trained model. Due to the information redundancy in the intermediate representations, both Head2Toe and VQT further adopt group lasso \cite{yuan2006model, argyriou2006multi} for feature selection. 

In contrast to previous works, we provide an insightful analysis that  entangled combination of task-specific
learning and pre-trained knowledge utilization results in high memory consumption of traditional PETL methods. Therefore, we propose SynQT to disentangle PETL into two distinct modules, with the aim of capturing task-specific knowledge and extracting useful features for downstream tasks, respectively. Such a disentangled paradigm enjoys memory efficiency in transfer learning compared with traditional PETL methods. 
\section{Method}

\subsection{Overall Framework}
As can be seen from Figures \ref{fig:framework} and \ref{fig:head}, SynQT makes the prediction relying on the features produced by stacked blocks, which take the intermediate representations of the original pre-trained model as external input.
In SynQT, each block mainly includes two parts: Query Synthesis Module (QSM) and Knowledge Extraction Module (KEM).
QSM is a lightweight module, which is used to capture task-specific knowledge, connecting the pre-trained model with the downstream task.
KEM uses the tokens generated by QSM as queries, and the intermediate features in the pre-trained ViT as keys and values. It reuses the original model's weights to extract features that are beneficial to downstream tasks.
Finally, the rich hierarchical features extracted by the KEMs are fed into the classification head, as shown in Figure \ref{fig:head}. We will elaborate our designs in the following sections.

\subsection{Query Synthesis Module}
\label{sec:qse}
PETL integrates the task-specific knowledge into the adapted model through training on the downstream dataset. Typically, existing methods \cite{jia2022visual,houlsby2019parameter,hu2021lora,chen2022adaptformer} insert additional trainable structures into the pre-trained model. However, this is the root of high memory consumption, since it changes all original features, and all intermediate features are modified and stored to be involved in calculating gradients. From the disentangled perspective, we solve this problem with a task-specific learning module, named Query Synthesis Module (QSM), which is trainable, lightweight, and structurally independent from the pre-trained model.


The purpose of QSM is to synthesize query tokens, interacting with intermediate representations of the original pre-trained model. 
Besides outputs from the previous block, we introduce a trainable prompt into the input space of each QSM.
These per-block additional prompts increase the capacity of QSM to capture task-specific knowledge. 
Inspired by the Transformer, QSM includes an attention layer and an FFN. Specifically, the input tokens interact with each other through a self-attention layer and are further fed into the feed-forward network for producing the query tokens. Besides, to keep the parameter efficiency, another design principle is to extensively use bottleneck-structured fully connected layers to reduce the number of parameters.

Let $n$ and $d$ denote the number of tokens and the hidden size in QSM, respectively. Note that $n$ is very small, $e.g.$, $n=4$, much lower than the number of tokens in the original model and the value of $d$ corresponds to that of the pre-trained model.
Before the self-attention layer, the output of the last block ${\hat{H}}_{i-1} \in \mathbb{R}^{n \times d}$ are added with trainable prompt $P_i$ and then fed into bottleneck projection layers.
It is formally formulated as follows,
\begin{equation}
    {Z_i'} = (({H}_{i} + P_i)\cdot \mathbf{W}_{down}) \cdot \mathbf{W}_{up},
\end{equation}
where $\mathbf{W}_{down}\in \mathbb{R}^{d \times \hat{d}}$ and $\mathbf{W}_{up}\in \mathbb{R}^{\hat{d} \times d}$ denote the down-projection and up-projection layer respectively, with $\hat{d} \ll d$. 

In the attention layer, $Z_i'$ is projected by three bottleneck modules to produce the query $Q^{qs}_i$, key $K^{qs}_i$, and value $V^{qs}_i$. Then
the hidden embedding after attention layers $Z_i''$ can be calculated as follows, controlled by a scale factor $s'$:  
\begin{equation}
    Z_i'' = s'\cdot\text{Attn}(Q^{qs}_i, K^{qs}_i, V^{qs}_i) + Z_i'.
\end{equation}
Then we can further obtain the synthesized query $\hat{H}_i$ after the FFN as the input of KEM, controlled by another scale factor $s''$:
\begin{equation}
    \hat{H}_i = s''\cdot\text{FFN}(\text{LN}(Z_i'')) + Z_i'',
\end{equation}
where FFN also consists of bottleneck-structured fully connected layers and $\text{LN}(\cdot)$ represents the layer normalization \cite{ba2016layer}. 
$s'$ and $s''$ are hyper-parameters to modulate the effects of the attention and FFN on the synthesized query.
Since QSM is trained on the downstream dataset from scratch, we believe that its output $\hat{H}_i$ captures the task-specific knowledge. We use $\hat{H}_i$ in a query-only manner as illustrated in the next section.

\subsection{Knowledge Extraction Module}
With the task-specific query obtained from QSM, we propose Knowledge Extraction Module (KEM) to exploit intermediate features of the original pre-trained model. Before that, we obtain original features $\{X_1, X_2, ..., X_l\}$ of the input sample from forward-propagation, where $l$ is the number of transformer blocks of the pre-train model. Note that this process is efficient due to no gradient calculation. 

KEM takes the synthesized query $\hat{H}_i$ and original features $\{X_1, X_2, ..., X_l\}$ as inputs and outputs useful features for the downstream task. Since $\hat{H}_i$ has a small number of tokens and original features are only queried by $\hat{H}_i$, there are no heavy features in the middle of KEM. Besides, we find that it is effective to reuse and freeze the pre-trained parameters for feature interaction in KEM. These designs ensure the parameter- and memory-efficiency of our method.

Specifically, with the projection weight in the pre-trained ViT blocks, we first obtain the query $Q^{ke}_i$, key $K^{ke}_i$, and value $V^{ke}_i$. Then follow-up attention calculation is performed by:
\begin{equation}
    E_i = \text{Attn}(Q^{ke}_i, K^{ke}_i, V^{ke}_i) + \hat{H}_i.
\end{equation}
We also reuse the weights of FFN to perform the same space transformation as the pre-trained model: 
\begin{equation}
    {H}_i = \text{FFN}(\text{Norm}(E_i)) + E_i ,
\end{equation}
where $\text{Norm}(\cdot)$ denotes the normalization used in the pre-trained model. 

To maximize the potential of KEM,  apart from the final output ${H}_i$, we additionally leverage features $F_i^{att}$ and $F_i^{ffn}$ after the attention module and the FFN for classification. They are obtained as follows:
\begin{equation}
    \left\{
    \begin{aligned}
    F_i^{att} &= \text{Attn}(Q^{ke}_i, K^{ke}_i, V^{ke}_i) \\
    F_i^{ffn} &= \text{FFN}(\text{Norm}(E_i))
    \end{aligned}.
    \right.
\end{equation}

\subsection{Classification Head}
\begin{wrapfigure}[25]{tr}{0.42\columnwidth}
\centering 
\includegraphics[width=0.42\textwidth]{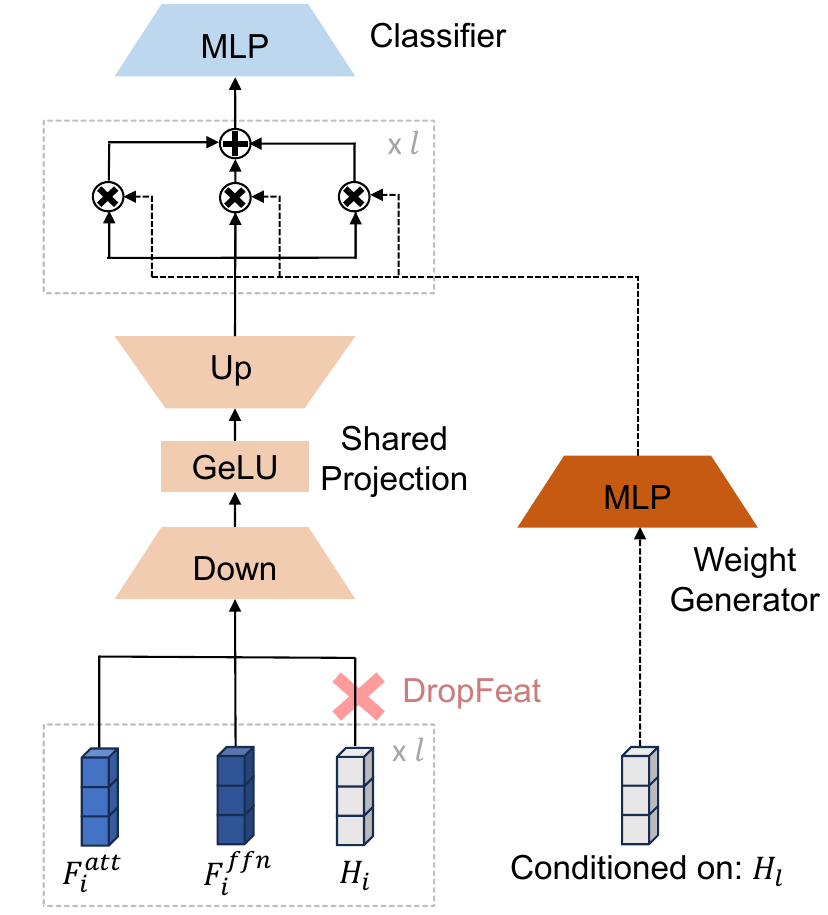}
\caption{\textbf{Classification head in SynQT}, corresponding to the head in Figure \ref{fig:framework}. We utilize a shared projection to align the hierarchical features and apply feature aggregation conditioned on the output of the last block $H_l$. We also adopt DropFeat to prevent overfitting, which randomly drops some features during training.}
\label{fig:head}
\end{wrapfigure}

The classification head makes a prediction of the input sample relying on rich and hierarchical features from KEMs, $\{F_1^{att}$, $F_1^{ffn}$, $H_1,...,F_l^{att}$, $F_l^{ffn}$, $H_l\}$, where $l$ is the number of blocks of the pre-trained ViT.
Instead of directly applying an MLP classifier on top of these features, we propose several techniques for better performance. Figure \ref{fig:head} shows the overall architecture of the classification head.

\noindent \textbf{Feature projection.}
Since these features are generated across different layers from KEMs, they may be misaligned in the representation space. Therefore, before aggregating all features, we align all features with projection layers. 
We use bottleneck-structured fully connected layers and share the projection layers across all features to improve the parameter efficiency.

\noindent \textbf{Conditional feature aggregation.}
These features contain a large amount of information but also some redundancy, perhaps hurting the performance of downstream tasks. Therefore, it is crucial to perform feature selection, aiming to extract the truly beneficial information. We propose a strategy of conditional feature aggregation, which generates weights conditioned on $H_l$ and performs a weighted-sum over all features. Our strategy enables SynQT to adaptively rely on different features to classify different samples. Despite only a few parameters, the conditional feature aggregation results in better performance compared to a simple average or weighted sum with learnable but fixed weights, as shown in our experiments.

\noindent \textbf{DropFeat in training.}
Due to rich features fed into the classification head, overfitting likely happens and degrades the performance. To solve this issue, inspired by the idea of dropout \cite{srivastava2014dropout,wan2013regularization,larsson2016fractalnet}, we propose DropFeat, a simple regularization technique to prevent overfitting. DropFeat randomly removes some input features for the classification head during training, while all features are leveraged during inference.
DropFeat ensures that the model does not overly rely on a specific feature, which significantly improves the model's performance.

\section{Experiment}
\subsection{Setup}
\textbf{Datasets.} We conduct experiments on the VTAB-1K benchmark and the fine-grained visual classification (FGVC) datasets.
VTAB-1K consists of 19 different classification tasks categorized into three groups: Natural, Specialized, and Structured. Each task of VTAB-1K contains 1000 training samples. 
We evaluate the few-shot performance of SynQT on 5 FGVC datasets: FGVC-Aircraft \cite{maji2013fine}, OxfordPets \cite{parkhi2012cats}, Food-101 \cite{bossard2014food}, Stanford Cars\cite{krause20133d}, and Oxford-Flowers102 \cite{nilsback2006visual}. We follow the split in \cite{zhang2022NOAH} and report the results in 1, 2, 4, 8, and 16-shot settings. The accuracy is averaged over three trials with different seeds.


\begin{table*}[t]
    \centering
    \caption{\textbf{Full results on VTAB-1K benchmark with pre-trained ViT-B/16 on ImageNet21K.} ``\# Memory'' specifies memory cost during training with 64 as batch size on CIFAR100. We report the Top-1 accuracy and the results are averaged over
group-wise mean values.}
    \label{tab:vtab}
    \setlength{\tabcolsep}{1pt}
    \resizebox{\textwidth}{!}{
        \begin{tabular}{p{2.5cm}<{}p{1.0cm}<{\centering}|
            p{0.75cm}<{\centering}p{0.75cm}<{\centering}p{0.75cm}<{\centering}p{0.75cm}<{\centering}p{0.75cm}<{\centering}p{0.75cm}<{\centering}p{0.75cm}<{\centering}|p{0.75cm}<{\centering}p{0.75cm}<{\centering}p{0.75cm}<{\centering}p{0.75cm}<{\centering}|p{0.75cm}<{\centering}p{0.75cm}<{\centering}p{0.75cm}<{\centering}p{0.75cm}<{\centering}p{0.75cm}<{\centering}p{0.75cm}<{\centering}p{0.75cm}<{\centering}p{0.75cm}<{\centering}|p{0.75cm}<{\centering}}
            \toprule[1.5pt]
            \multicolumn{2}{c|}{}             & \multicolumn{7}{c|}{\textbf{Natural}}                          & \multicolumn{4}{c|}{\textbf{Specialized}} & \multicolumn{8}{c|}{\textbf{Structured}}                                                                                                                               \\
                                              & \multicolumn{1}{c|}{{\rotatebox[origin=c]{90}{\# Memory (GB)}}}
                                              & \multicolumn{1}{c}{{\rotatebox[origin=c]{90}{CIFAR100}}}
                                              & \multicolumn{1}{c}{{\rotatebox[origin=c]{90}{Caltech101}}}
                                              & \multicolumn{1}{c}{{\rotatebox[origin=c]{90}{DTD}}}
                                              & \multicolumn{1}{c}{{\rotatebox[origin=c]{90}{Flower102}}}
                                              & \multicolumn{1}{c}{{\rotatebox[origin=c]{90}{Pets}}}
                                              & \multicolumn{1}{c}{{\rotatebox[origin=c]{90}{SVHN}}}
                                              & \multicolumn{1}{c|}{{\rotatebox[origin=c]{90}{Sun397}}}
                                              & \multicolumn{1}{c}{{\rotatebox[origin=c]{90}{Camelyon}}}
                                              & \multicolumn{1}{c}{{\rotatebox[origin=c]{90}{EuroSAT}}}
                                              & \multicolumn{1}{c}{{\rotatebox[origin=c]{90}{Resisc45}}}
                                              & \multicolumn{1}{c|}{{\rotatebox[origin=c]{90}{Retinopathy}}}
                                              & \multicolumn{1}{c}{{\rotatebox[origin=c]{90}{Clevr-Count}}}
                                              & \multicolumn{1}{c}{{\rotatebox[origin=c]{90}{Clevr-Dist}}}
                                              & \multicolumn{1}{c}{{\rotatebox[origin=c]{90}{DMLab}}}
                                              & \multicolumn{1}{c}{{\rotatebox[origin=c]{90}{KITTI-Dist}}}
                                              & \multicolumn{1}{c}{{\rotatebox[origin=c]{90}{dSpr-Loc}}}
                                              & \multicolumn{1}{c}{{\rotatebox[origin=c]{90}{dSpr-Ori}}}
                                              & \multicolumn{1}{c}{{\rotatebox[origin=c]{90}{sNORB-Azim}}}
                                              & \multicolumn{1}{c|}{{\rotatebox[origin=c]{90}{sNORB-Ele}}}
                                              & \multicolumn{1}{c}{{\rotatebox[origin=c]{90}{Average}}}                                                                                                                                                                                                                             \\
            \specialrule{0em}{1pt}{1pt}
            \hline
            \specialrule{0em}{1pt}{1pt}
            \multicolumn{22}{l}{\emph{Traditional Fine-Tuning}}                                                                                                                                                                                                                                                                      \\
            \hline
            \specialrule{0em}{1pt}{1pt}

            Linear Probing                    &  2.5                                                              & 50.6                                      & 85.6                                     & 61.4 & 79.5 & 86.5 & 40.8 & 38.0 & 79.7 & 91.5 & 71.7 & 65.5 & 41.4 & 34.4 & 34.1 & 55.4 & 18.1 & 26.4 & 16.5 & 24.8 & 52.7 \\
            Full Fine-Tuning                       &  11.1                                                               & 44.3                                      & 84.5                                     & 54.1 & 84.7 & 74.7 & \textbf{87.2} & 26.9 & \textbf{85.3} & 95.0 & 76.0 & 70.4 & 71.5 & 60.5 & 46.9 & 72.9 & \textbf{74.5} & 38.7 & 28.5 & 23.8 & 63.2 \\
            \specialrule{0em}{1pt}{1pt}
            \hline
            \specialrule{0em}{1pt}{1pt}
            \multicolumn{22}{l}{\emph{PETL Methods}}                                                                                                                                                                                                                                                                                \\
            \specialrule{0em}{1pt}{1pt}
            \hline
            \specialrule{0em}{1pt}{1pt}
            BitFit   \cite{zaken2021bitfit}                         & 7.8                                                          & 72.8                                      & 87.0                                     & 59.2 & 97.5 & 85.3 & 59.9 & \textbf{51.4} & 78.7 & 91.6 & 72.9 & 69.8 & 61.5 & 55.6 & 32.4 & 55.9 & 66.6 & 40.0 & 15.7 & 25.1 & 65.2 \\
            VPT-Shallow  \cite{jia2022visual}                     & 8.1                                                          & 77.7                                      & 86.9                                     & 62.6 & 97.5 & 87.3 & 74.5 & 51.2 & 78.2 & 92.0 & 75.6 & 72.9 & 50.5 & 58.6 & 40.5 & 67.1 & 68.7 & 36.1 & 20.2 & 34.1 & 67.8 \\
            VPT-Deep  \cite{jia2022visual}                        & 8.1                                                          & \textbf{78.8}                                      & \textbf{90.8}                                     & 65.8 & 98.0 & 88.3 & 78.1 & 49.6 & 81.8 & 96.1 & 83.4 & 68.4 & 68.5 & 60.0 & 46.5 & 72.8 & 73.6 & 47.9 & \textbf{32.9} & \textbf{37.8} & 72.0 \\
            \hline
            \specialrule{0em}{1pt}{1pt}
            \multicolumn{22}{l}{\emph{Memory-Efficient PETL Methods}}                                                                                                                                                                                                                                                                     \\
            \hline
            \specialrule{0em}{1pt}{1pt}

            VQT  \cite{tu2023visual}                             & 3.2                                                          & 66.3                                      & 89.9                                     & 67.8 & 97.9 & 84.7 & 79.9 & 45.5 & 79.0 & 95.2 & 80.9 & 74.7 & 46.7 & 61.6 & 45.1 & 63.6 & 62.9 & 32.1 & 30.0 & 28.8 & 68.3 \\

            \hline
            \specialrule{0em}{1pt}{1pt}

            \rowcolor{lightgray}\textbf{SynQT (Ours)} & 3.4                                                 & 70.9                                      & 89.7                                     & \textbf{68.8} & \textbf{98.5} & \textbf{89.6} & 77.8 & 50.6 & 82.3 & \textbf{96.7} & \textbf{83.5} & \textbf{75.2} & \textbf{71.8} & \textbf{62.7} & \textbf{48.5} & \textbf{75.4} & 74.1 & \textbf{49.0} & 31.7 & 36.1 & \textbf{72.9}
            \\
            \bottomrule[1.5pt]
        \end{tabular}
    }
\end{table*}
\noindent\textbf{Implementation details.}
We utilize a pre-trained ViT-B/16 \cite{dosovitskiy2020image} on the ImageNet-21K dataset as the foundational model. We also conduct experiments on the ImageNet-1K supervised pre-trained backbone, as well as on the self-supervised checkpoints (DINO \cite{caron2021emerging} and MAE \cite{he2022masked}). 
For a lightweight design of QSM, we use an extremely small number of tokens, with the number of tokens $n$ from $\{1, 4, 16\}$. This is significantly fewer than the number of tokens in VPT, which can reach up to 200. 
Additionally, the hyper-parameters $s'$ and $s''$  are chosen from $\{0.1, 1\}$. 
The hidden size of the bottleneck projection is set to 48, except attention projection of the query, key, and value in QSM, which is set to 8.

Due to the limited number of validation samples in VTAB-1K, we implement 5-fold cross-validation on the training sets to determine the optimal hyper-parameters. For few-shot learning, we use the default splits of FGVC datasets to search for the best hyperparameters. 
The Top-1 accuracy is reported on the test sets in all our experiments.
Following common practice \cite{jie2023fact}, we adopt AdamW \cite{loshchilov2017decoupled} as the optimizer and a cosine learning rate scheduler with a constant batch size of 64. 
For data augmentation, we adjust the image size to 224x224 directly for VTAB-1K, and for few-shot learning, we resize the image to 256x256 before applying a center crop.

\begin{table}[t]
    \caption{\textbf{Classification results on MAE, DINO, and Supervised ImageNet-1K pre-trained backbones. } We report the Top-1 accuracy and the averaged results.}
    \label{tab:backbone}
    \setlength{\tabcolsep}{10pt}
    \scriptsize
    \centering
    \resizebox{0.85\textwidth}{!}{
    \begin{tabular}{lcccc}
        \toprule
        Methods          & Natural         & Specialized              & Structured               & Average                  \\
        \midrule
        \multicolumn{5}{c}{MAE backbone}                                                                                    \\
        \midrule
        BitFit \cite{zaken2021bitfit}          & 63.5            & 80.6                     & 57.2                     & 67.1                     \\
        VPT-Shallow  \cite{jia2022visual}    & 38.3            & 72.0                     & 25.0                       & 45.1                     \\
        VPT-Deep  \cite{jia2022visual}       & 50.8            & 76.4                     & 37.3                     & 54.8                     \\
        VQT  \cite{tu2023visual}            & 56.6            & 78.6                     & 43.4                     & 59.5                     \\
        \baseline{\textbf{SynQT (Ours)}} & \baseline{\textbf{66.0}} & \baseline{\textbf{82.6}}          & \baseline{\textbf{58.2}}          & \baseline{\textbf{68.9}}         \\
        \midrule
        \multicolumn{5}{c}{DINO backbone}                                                                                   \\
        \midrule
        BitFit  \cite{zaken2021bitfit}         & \textbf{76.8}            & 83.8                     & 49.2                     & 69.9                     \\
        VPT-Shallow  \cite{jia2022visual}    & 69.8            & 82.6                     & 38.5                     & 63.6                     \\
        VPT-Deep  \cite{jia2022visual}       & 76.7            & 83.8                     & 47.3                     & 69.3                     \\
        VQT  \cite{tu2023visual}            & 71.0            & 83.6                     & 37.2                     & 63.9                     \\
        \baseline{\textbf{SynQT (Ours)}} & \baseline{75.1} & \baseline{\textbf{84.8}}          & \baseline{\textbf{56.9}}          & \baseline{\textbf{72.3}}          \\
        \midrule
        \multicolumn{5}{c}{Supervised ImageNet-1K backbone}                                                                 \\
        \midrule
        BitFit \cite{zaken2021bitfit}          & 74.6            & 83.4                     & 49.4                     & 69.1                     \\
        VPT-Shallow \cite{jia2022visual}     & 63.7            & 80.3                     & 40.2                     & 61.4                     \\

        VPT-Deep \cite{jia2022visual}        & \textbf{74.9}            & 82.9                     & 53.9                     & 70.6                     \\
        VQT    \cite{tu2023visual}          & 72.7            & 84.5                     & 49.3                     & 68.8                     \\
        \baseline{\textbf{SynQT (Ours)}} & \baseline{73.2} & \baseline{\textbf{84.7}} & \baseline{\textbf{56.2}} & \baseline{\textbf{71.4}} \\
        \bottomrule
    \end{tabular}}
\end{table}

\subsection{Main Results} 
Besides the traditional fine-tuning methods (linear probing and full fine-tuning), we compare our method with PETL methods including BitFit \cite{zaken2021bitfit}, VPT-Shallow \cite{jia2022visual}, and VPT-Deep \cite{jia2022visual}, as well as a memory-efficient one, VQT \cite{tu2023visual}, on VTAB-1K. 
As can be seen from Table \ref{tab:vtab}, the most striking observation emerging from
the comparison is that SynQT outperforms other methods while saving up to 50\% of GPU memory. 
Given the same batch size, previous PETL methods rely on the full backward propagation of the original model and consume 7$\sim$8 GB of memory, whereas SynQT disentangles transfer learning and reduces the memory cost to 3.4 GB. 
Notably, our method performs better than VPT-Deep on 14 of 19 tasks. 
We notice that although the memory cost of SynQT is slightly higher, it achieves a significant performance improvement of 4.6\% compared to VQT.

In order to further validate the robustness of our method, we compare SynQT with baseline models across various checkpoints which are derived from different    
learning paradigms, including supervised learning (pretrained over ImageNet-
\begin{wrapfigure}[13]{tr}{0.41\columnwidth}
        \centering
    \includegraphics[width=\linewidth]{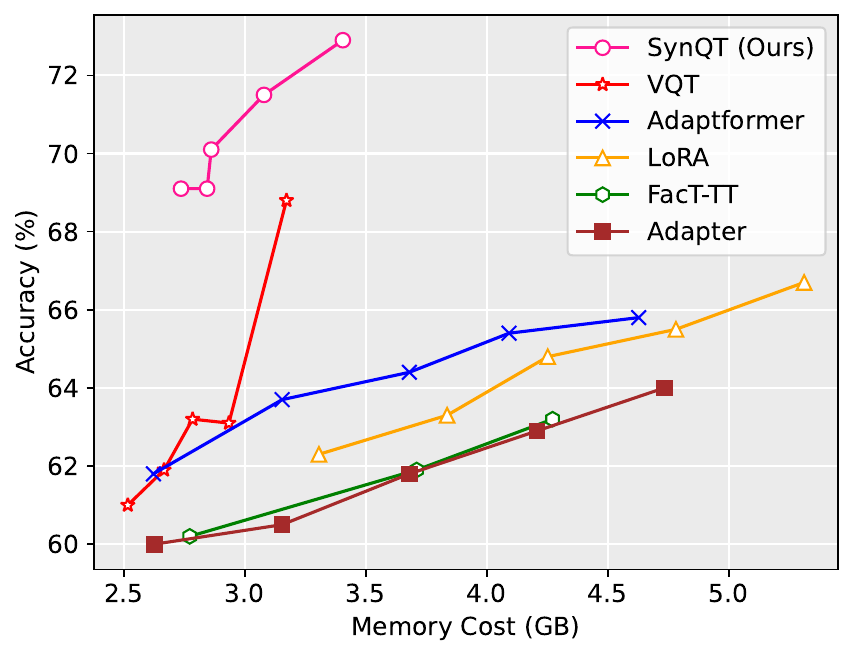}
    \caption{\textbf{Performance comparison under memory constraints.} SynQT outperforms other methods with different memory costs.}
    \label{fig:memory}
\end{wrapfigure}
21K \cite{deng2009imagenet}), 
and self-supervised learning (DINO \cite{caron2021emerging} and MAE \cite{he2022masked}).
The results in Table \ref{tab:backbone} indicate that SynQT shows strong compatibility and outperforms  PETL methods by a substantial margin over different backbones. 
For instance, compared to the strongest baselines, our method achieves improvements of 1.8\%, 3.3\%, and 0.8\% on average with MAE, DINO, and supervised ImageNet-1K backbones, respectively.

\subsection{Memory-Efficiency Evaluation}
Previous PETL methods \cite{hu2021lora, chen2022adaptformer, houlsby2019parameter, jie2023fact}  focus on reducing the number of 
parameters when fine-tuning the model. While decreasing the cost of model storage, these methods are not widely applicable in practical training because they still require a significant amount of GPU memory. Therefore, we propose SynQT to overcome this problem. In this  section, we highlight our improvements in terms of memory costs by comparing our method with some SOTA PETL methods under a given GPU memory constraint.

\begin{table*}[t]
    \vspace{-0.1in}
    \centering
    \caption{\textbf{Full results on VTAB-1K benchmark under memory constraints}. The memory is constrained within 5GB and ``\# Memory'' specifies the memory cost during training with 64 as batch size on CIFAR100. We report the Top-1 accuracy and the results are averaged over
group-wise mean values.}
    \label{tab:vtab-memory}
    \setlength{\tabcolsep}{1pt}
    \resizebox{\textwidth}{!}{
        \begin{tabular}{p{2.5cm}<{}p{1.0cm}<{\centering}|
            p{0.75cm}<{\centering}p{0.75cm}<{\centering}p{0.75cm}<{\centering}p{0.75cm}<{\centering}p{0.75cm}<{\centering}p{0.75cm}<{\centering}p{0.75cm}<{\centering}|p{0.75cm}<{\centering}p{0.75cm}<{\centering}p{0.75cm}<{\centering}p{0.75cm}<{\centering}|p{0.75cm}<{\centering}p{0.75cm}<{\centering}p{0.75cm}<{\centering}p{0.75cm}<{\centering}p{0.75cm}<{\centering}p{0.75cm}<{\centering}p{0.75cm}<{\centering}p{0.75cm}<{\centering}|p{0.75cm}<{\centering}}
            \toprule[1.5pt]
            \multicolumn{2}{c|}{}             & \multicolumn{7}{c|}{\textbf{Natural}}                          & \multicolumn{4}{c|}{\textbf{Specialized}} & \multicolumn{8}{c|}{\textbf{Structured}}                                                                                                                               \\
                                              & \multicolumn{1}{c|}{{\rotatebox[origin=c]{90}{ \# Memory (GB)}}}
                                              & \multicolumn{1}{c}{{\rotatebox[origin=c]{90}{CIFAR100}}}
                                              & \multicolumn{1}{c}{{\rotatebox[origin=c]{90}{Caltech101}}}
                                              & \multicolumn{1}{c}{{\rotatebox[origin=c]{90}{DTD}}}
                                              & \multicolumn{1}{c}{{\rotatebox[origin=c]{90}{Flower102}}}
                                              & \multicolumn{1}{c}{{\rotatebox[origin=c]{90}{Pets}}}
                                              & \multicolumn{1}{c}{{\rotatebox[origin=c]{90}{SVHN}}}
                                              & \multicolumn{1}{c|}{{\rotatebox[origin=c]{90}{Sun397}}}
                                              & \multicolumn{1}{c}{{\rotatebox[origin=c]{90}{Camelyon}}}
                                              & \multicolumn{1}{c}{{\rotatebox[origin=c]{90}{EuroSAT}}}
                                              & \multicolumn{1}{c}{{\rotatebox[origin=c]{90}{Resisc45}}}
                                              & \multicolumn{1}{c|}{{\rotatebox[origin=c]{90}{Retinopathy}}}
                                              & \multicolumn{1}{c}{{\rotatebox[origin=c]{90}{Clevr-Count}}}
                                              & \multicolumn{1}{c}{{\rotatebox[origin=c]{90}{Clevr-Dist}}}
                                              & \multicolumn{1}{c}{{\rotatebox[origin=c]{90}{DMLab}}}
                                              & \multicolumn{1}{c}{{\rotatebox[origin=c]{90}{KITTI-Dist}}}
                                              & \multicolumn{1}{c}{{\rotatebox[origin=c]{90}{dSpr-Loc}}}
                                              & \multicolumn{1}{c}{{\rotatebox[origin=c]{90}{dSpr-Ori}}}
                                              & \multicolumn{1}{c}{{\rotatebox[origin=c]{90}{sNORB-Azim}}}
                                              & \multicolumn{1}{c|}{{\rotatebox[origin=c]{90}{sNORB-Ele}}}
                                              & \multicolumn{1}{c}{{\rotatebox[origin=c]{90}{Average}}}                                                                                                                                                                                                                             \\
            \hline
            \specialrule{0em}{1pt}{1pt}
            FacT-TT  \cite{jie2023fact}                         & 4.3                                                          & 40.4                                      & 84.2                                     & 59.8 & 96.8 & 86.6 & 53.4 & 45.8 & 79.4 & 92.2 & 73.3 & 74.4 & 59.2 & 50.2 & 42.5 & 68.8 & 48.2 & 25.9 & 19.8 & 28.7 & 63.2 \\
            Adapter  \cite{houlsby2019parameter}                         & 4.7                                                         & 64.9                                      & 86.0                                     & 62.8 & 97.3 & 86.6 & 46.8 & 44.4 & 79.8 & 93.9 & 76.7 & 74.6 & 63.3 & 40.2 & 38.9 & 68.5 & 33.1 & 32.3 & 19.8 & 30.5 & 64.0 \\
            LoRA     \cite{hu2021lora}                         & 4.8                                                          & 62.5                                      & 82.1                                     & 60.4 & 97.5 & 86.6 & 58.6 & 45.3 & 80.6 & 95.3 & 78.8 & 72.8 & 64.8 & 45.1 & 41.3 & 68.5 & 49.6 & 28.1 & 24.6 & 30.5 & 65.5 \\
            AdaptFormer \cite{chen2022adaptformer}                      & 4.6                                                          & 65.6                                      & 86.1                                     & 62.0 & 96.9 & 86.2 & 35.5 & 45.6 & \textbf{82.6} & 96.2 & 77.6 & 74.4 & 66.2 & 47.7 & 44.3 & 72.3 & 55.6 & 29.4 & 22.6 & 33.9 & 65.8 \\

            \rowcolor{lightgray}\textbf{SynQT(Ours)} & \textbf{3.4 }                                                & \textbf{70.9}                                      & \textbf{89.7}                                     & \textbf{68.8} & \textbf{98.5} & \textbf{89.6} & \textbf{77.8} & \textbf{50.6} & 82.3 & \textbf{96.7} & \textbf{83.5} & \textbf{75.2} & \textbf{71.8} & \textbf{62.7} & \textbf{48.5} & \textbf{75.4} & \textbf{74.1} & \textbf{49.0} & \textbf{31.7} & \textbf{36.1} & \textbf{72.9}
            \\
            \bottomrule[1.5pt]
        \end{tabular}
    }
\end{table*}


For baseline methods, we discard some specific trainable structures to meet the memory constraint.
Specifically, we remove these additional structures near the input layer, which means that the training only requires a partial back-propagation with a lower memory usage.
We progressively remove the structure nearest to the input layer until it meets the given memory constraint. Besides, for a fair comparison, we reduce the memory usage of SynQT in the same manner.

We compare SynQT with SOTA baselines including FacT-TT \cite{jie2023fact}, Adapter \cite{houlsby2019parameter}, LoRA \cite{hu2021lora} and AdaptFormer \cite{chen2022adaptformer} under such a constraint and report the results in Table \ref{tab:vtab-memory}. 
While previous methods can achieve promising performance with few trainable parameters, they unfortunately suffer from significant performance drops when such memory constraints are imposed.
In contrast, SynQT excels in this scenario, achieving state-of-the-art performance and surpassing AdaptFormer by 7.1\% and VQT by 4.6\%. 

To have a better understanding of how memory constraints affect model performance, we estimate the results of the methods mentioned above across varying memory costs. As depicted in Figure \ref{fig:memory}, SynQT continues to outperform the baselines across different memory costs. Notably, when the constraint is very strict, $i.e.$, 3 GB, most of the existing PETL methods perform poorly, with a gap of almost 7\% compared to SynQT.
Moreover, it is noteworthy that relaxing the memory constraints improves the performance of all the methods, especially for a significant improvement observed for our SynQT.


\subsection{Few-Shot Learning on FGVC}
\begin{figure*}[tbp]
    \centering
    \includegraphics[width=0.99\textwidth]{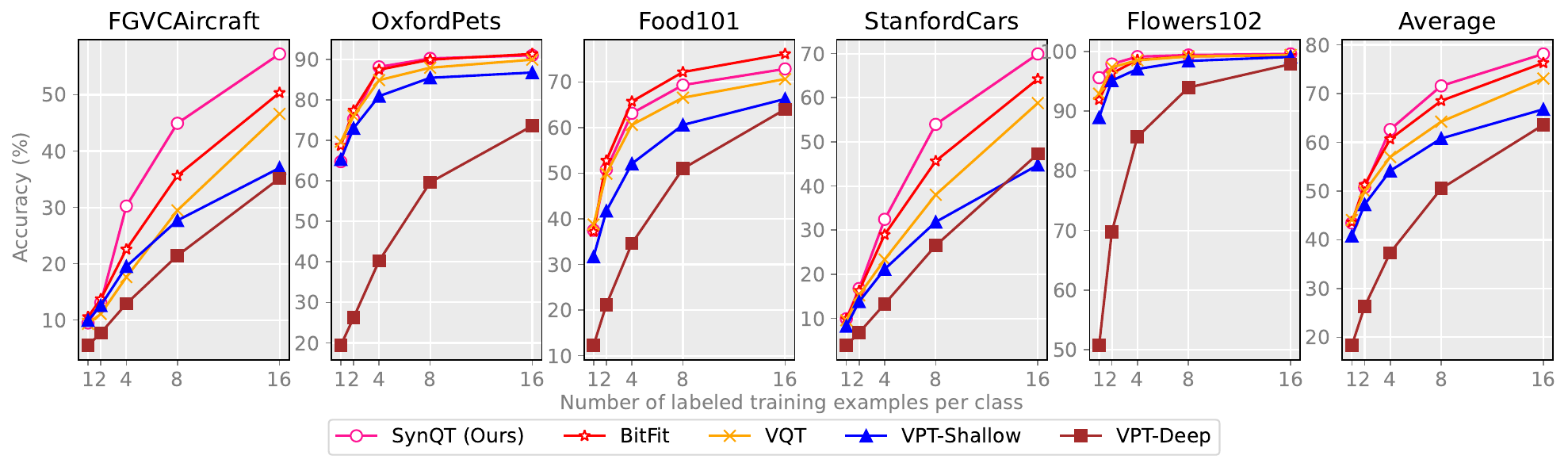}
    \caption{\textbf{Performance comparison in few-shot learning.} Overall, SynQT outperforms other baselines in the extreme low-data regime. }
    \label{fig:few-shot}
\end{figure*}

To further evaluate SynQT's ability to transfer knowledge from the pre-trained model using few samples, we compare SynQT with BitFit \cite{zaken2021bitfit}, VPT-Shallow \cite{jia2022visual}, VPT-Deep \cite{jia2022visual}, and VQT \cite{tu2023visual} in few-shot setting on 5 FGVC datasets.
The results are reported in Figure \ref{fig:few-shot}.   

Overall, SynQT significantly outperforms all other baselines in 4, 6, 8, and 16-shot scenarios, and its performance is comparable in the 1-shot scenario. The gaps between VQT and SynQT indicate that the lightweight, task-specific QSM enables SynQT to capture beneficial information from samples in the extremely low-data regime. Simply incorporating a trainable prompt as a query, as in VQT, is not sufficient to handle such challenging few-shot tasks.
Interestingly, VPT-Shallow outperforms VPT-Deep in the low-data regime, due to the fact that the larger number of prompts in VPT-Deep are not well-trained on the limited data. This suggests that merely increasing the number of prompt numbers could potentially lead to a performance drop.

\begin{table}[t]
\centering 
\caption{\textbf{Applying SynQT to Swin-B and ConvNeXt-B.} We report the Top-1 accuracy and the averaged results.}
\label{tab:extension}
\tablestyle{5pt}{1.05}
\resizebox{1\columnwidth}{!}{
\begin{tabular}{l|cccc|cccc}
\toprule
 \multirow{2}{*}{Method} &  \multicolumn{4}{c|}{Swin-B} & \multicolumn{4}{c}{ConvNeXt-B}   \\
 \cline{2-9}
& Natural & Specialized & Structured & Average & Natural & Specialized & Structured & Average \\
\cline{1-9}
         Linear           &73.5& 80.8 &33.5 &62.6 &74.5 & 81.5 & 34.8 & 63.6                 \\
        VPT-Deep & 76.8 &84.5 &53.4&71.6 & \textbf{78.5} &\textbf{83.0}& 44.6&68.7 \\
        \textbf{SynQT} & \textbf{77.5} & \textbf{86.0} & \textbf{57.8} & \textbf{73.8}  & 78.1 & 82.1 & \textbf{48.2} & \textbf{69.5}        \\
\bottomrule
\end{tabular}
}
\end{table}

\subsection{Versatility of SynQT}
To demonstrate the versatility of SynQT, we apply SynQT to different model architectures and combine SynQT with existing PETL methods.

\noindent \textbf{Applying SynQT to different model architectures.}
Here, we apply SynQT to an advanced Transformer-like architecture (Swin \cite{liu2021swin}) and a convolutional neural network (ConvNeXt \cite{liu2022convnet}). 
Following \cite{jia2022visual}, we introduce prompts for each window in Swin and incorporate learnable prompt pixels for the feature map in ConvNeXt. 
Specifically, due to the varying hidden sizes and number of blocks in different layers, we extract features from the third layer of Swin and the last layer of ConvNext to ensure sufficient extraction of beneficial information.
We find that both of them significantly benefit from SynQT, 
as shown in Table \ref{tab:extension}. On average, SynQT outperforms VPT-Deep by 2.2\% and 1.2\% for Swin and ConvNeXt, respectively.

\noindent \textbf{Combining SynQT with existing PETL methods.}
We further jointly utilize SynQT and existing PETL methods to adapt pre-trained models, as shown in Table \ref{tab:versatility}. The results demonstrate that both VPT and AdaptFormer benefit from SynQT on VTAB-1K, with improvements of 4.4\% and 1.4\% on average, respectively.
We can see that for AdaptFormer and VPT, VQT may result in a performance drop in some cases (e.g., the Natural group), whereas equipping SynQT leads to significant performance improvements across all groups.

\begin{figure}[t]
\begin{minipage}{0.52\textwidth}

\captionof{table}{\textbf{Combining different PETL methods.} AF is short for AdaptFormer.}
\centering 
\resizebox{\linewidth}{!}{
\begin{tabular}{lcccc}
        \toprule
                Methods          & \multicolumn{1}{c}{Natural}         & \multicolumn{1}{c}{Specialized}             & \multicolumn{1}{c}{Structured}   & \multicolumn{1}{c}{Average}               \\
        \midrule                   
        AF          & 80.1 &82.3 &50.3 &70.9     \\
        AF+VQT       & 79.6   &84.3 &53.0 & 72.3        \\
        AF+SynQT          & \textbf{82.6} &  \textbf{85.4}  & \textbf{57.9} & \textbf{75.3}    \\
\midrule 
VPT             & 79.1 &84.6 &54.4 & 72.7 \\
VPT+VQT         & 78.9 & 83.7 & 54.6 &72.4\\
VPT+SynQT         & \textbf{81.1}&\textbf{84.9} &\textbf{57.7} &  \textbf{74.1}\\
        \bottomrule
\end{tabular}
}
\label{tab:versatility}
\end{minipage}
\begin{minipage}{0.47\textwidth}

\centering
\captionof{table}{\textbf{Necessity of disentanglement (KEM and QSM).}}
\label{tab:component}
    \centering
\resizebox{\columnwidth}{!}{
    \begin{tabular}{lcccc}
        \toprule
        Method          & Natural         & Specialized             & Structured   & Average                                                                                             \\
        \midrule
        \space Linear           & 63.2&
77.1&
31.4&
57.2                                             \\
        +KEM & 71.8&
80.1&
32.4&
61.4\\
        +QSM &77.8& 83.6& 55.0 & 72.1  \\
        +Head (SynQT) & \textbf{78.0} & \textbf{88.4} & \textbf{56.2} & \textbf{72.9} \\
        \bottomrule
    \end{tabular}
}

\end{minipage}
\end{figure}

\subsection{Ablation Study}

\begin{table*}[t]
    \caption{\textbf{SynQT ablation experiments} with ViT-B/16 with ImageNet-21K backbone.
        Default settings are marked in \colorbox{baselinecolor}{gray}.}
    \label{tab:ablations}
    \centering
    \subfloat[
        \textbf{Hidden size in QSM.} 48 as the width is a trade-off between parameters and performance.
        \label{tab:hidden}
    ]{
        \centering
        \begin{minipage}{0.45\linewidth}{\begin{center}
                    \tablestyle{5pt}{1.05}
                    \resizebox{\linewidth}{!}{
                    \begin{tabular}{x{16}x{32}x{16}x{16}x{16}x{16}}
                        Dim           & Param             & Nat.                     & Spe.            & Str.                     & Avg.                     \\
                        \shline
                        16            & 2.14 M            & 77.7                     & 83.8            & 55.8                     & 72.4                     \\
                        32            & 2.43 M            & 77.7                     & 83.9            & 55.3                     & 72.3                     \\
                        \baseline{48} & \baseline{2.73 M} & \baseline{\textbf{78.0}} & \baseline{84.4} & \baseline{\textbf{56.2}} & \baseline{\textbf{72.9}} \\
                        64            & 3.02 M            & 77.6                     & 84.4            & 56.1                     & 72.7                     \\
                        96            & 3.61 M            & 77.7                     & \textbf{84.5}   & 55.9                     & 72.7                     \\
                    \end{tabular}}
                \end{center}}\end{minipage}

    }
    \quad
        \subfloat[
        \textbf{Feature selection.}
        Hierarchical features contains rich information that benefit to downstream tasks.
        \label{tab:feat}
    ]{
        \begin{minipage}{0.45\linewidth}{\begin{center}
                    \tablestyle{7pt}{1.05}
                    \resizebox{\linewidth}{!}{
                    \begin{tabular}{x{48}x{16}x{16}x{16}x{16}}
                        SynQT             & Nat.                     & Spe.            & Str.                     & Avg.                     \\
                        \shline
                        w/o ${H}_i$       & 78.3                     & 84.0            & 53.8                     & 72.0                     \\
                        w/o  $F^{attn}_i$ & 78.3                     & 83.5            & 55.4                     & 72.4                     \\
                        w/o $F^{ffn}_i$   & 77.8                     & \textbf{84.7}   & 55.4                     & 72.6                     \\
                        \baseline{Full}   & \baseline{\textbf{78.0}} & \baseline{84.4} & \baseline{\textbf{56.2}} & \baseline{\textbf{72.9}} \\
                        \multicolumn{5}{c}{~}                                                                                                \\
                    \end{tabular}}
                \end{center}}\end{minipage}
    }
    \\
    \quad
    \subfloat[
        \textbf{Input of QSM. }Both the trainable prompt and the output from last block are important for generating task-specific query.
        \label{tab:synthesize}
    ]{
        \centering
        \begin{minipage}{0.45\linewidth}{\begin{center}
                    \tablestyle{5pt}{1.05}
                    \resizebox{\linewidth}{!}{
                    \begin{tabular}{x{69}x{16}x{16}x{16}x{16}}
                        SynQT           & Nat.                     & Spe.                     & Str.                     & Avg.                     \\
                        \shline
                        w/o Prompt      & 77.2                     & 82.9                     & 51.9                     & 70.7                     \\
                        w/o Last Output & 77.5                     & \textbf{84.4}            & 52.3                     & 71.4                     \\
                        \baseline{Full} & \baseline{\textbf{78.0}} & \baseline{\textbf{84.4}} & \baseline{\textbf{56.2}} & \baseline{\textbf{72.9}} \\
                    \end{tabular}}
                \end{center}}\end{minipage}

    }
    \quad
    \centering
    \subfloat[
        \textbf{Projection in the classification head.} Shared projection works the best for downstream tasks.
        \label{tab:projection}
    ]{
        \centering
        \begin{minipage}{0.45\linewidth}{\begin{center}
                    \tablestyle{6pt}{1.05}
                    \resizebox{\linewidth}{!}{
            \begin{tabular}{x{48}x{16}x{16}x{16}x{16}}
                Projection        & Nat.            & Spe.                     & Str.                     & Avg.                     \\
                \shline
                w/o Proj.         & \textbf{78.4}   & 83.8                     & 55.0                     & 72.4                     \\
                \baseline{Shared} & \baseline{78.0} & \baseline{\textbf{84.4}} & \baseline{\textbf{56.2}} & \baseline{\textbf{72.9}} \\
                Independent       & 77.9            & \textbf{84.4}            & 55.5                     & 72.6                     \\
            \end{tabular}}
        \end{center}}\end{minipage}
    }
    \\
    \subfloat[
        \textbf{Feature aggregation.}  Conditional feature aggregation is more effective than mean and learnable but fixed weight.
        \label{tab:aggregation}
    ]{
        \centering
        \begin{minipage}{0.45\linewidth}{\begin{center}
                    \tablestyle{5pt}{1.05}
                    \resizebox{\linewidth}{!}{
            \begin{tabular}{x{64}x{16}x{16}x{16}x{16}}
                SynQT                  & Nat.                     & Spe.            & Str.                     & Avg.                     \\
                \shline
                Simply Avg.            & 77.9                     & 84.6            & 55.6                     & 72.7                     \\
                Fixed Weights          & 77.9                     & \textbf{84.7}   & 55.5                     & 72.7                     \\
                \baseline{Conditional} & \baseline{\textbf{78.0}} & \baseline{84.4} & \baseline{\textbf{56.2}} & \baseline{\textbf{72.9}} \\
                
            \end{tabular}}
        \end{center}}\end{minipage}

    }
    \quad
    \subfloat[
        \textbf{DropFeat in the classification head.} The DropFeat module in the conditional head makes model more robust.
        \label{tab:dropfeat}
    ]{
        \centering
        \begin{minipage}{0.45\linewidth}{\begin{center}
                    \tablestyle{5pt}{1.05}
                    \resizebox{\linewidth}{!}{
            \begin{tabular}{x{64}x{16}x{16}x{16}x{16}}
                SynQT                  & Nat.                     & Spe.                     & Str.                     & Avg.                     \\
                \shline
                w/o DropFeat           & 77.7                     & 84.3                     & 55.2                     & 72.4                     \\
                \baseline{w/ DropFeat} & \baseline{\textbf{78.0}} & \baseline{\textbf{84.4}} & \baseline{\textbf{56.2}} & \baseline{\textbf{72.9}} \\
                \multicolumn{5}{c}{~}                                                                                                              \\
            \end{tabular}}
        \end{center}}\end{minipage}

    }
\end{table*}
We ablate our SynQT with different settings in Tables \ref{tab:component} and \ref{tab:ablations} for a better understanding of  SynQT. 

\noindent\textbf{Disentanglement.} To demonstrate the necessity of disentanglement, we perform an ablation in Table \ref{tab:component}. We set the baseline as linear probing since KEM exploits intermediate features. To gain deeper insights into the roles of QSM and Head, we employ random prompts to query in KEM as the model without QSM, and we utilize a simple average for aggregation as the model without Head.  The results demonstrate that our idea of disentangling (QSM and KEM) is the key to achieving superior performance. 

\noindent\textbf{Hidden size in QSM.}
We vary the hidden size of the bottleneck module in QSM and report the results in Table \ref{tab:hidden}. 
With a larger hidden size, the performance improves due to the increasing capacity. However, when the hidden size exceeds 48, the benefit of introducing new parameters diminishes and overfitting may happen in datasets with few training samples. The results indicate that redundant parameters might not enhance the model's performance, and a hidden size of 48 is a reasonable choice.

\noindent\textbf{Input of QSM.}
Table \ref{tab:synthesize} studies the influence of different inputs of QSM. We can see that
both the trainable prompts and the output from the last block are indispensable for QSM to generate high-quality task-specific query. The trainable prompts improve the capacity of QSM, and the results from the previous block ensure that knowledge is retained during the forward process. 

\noindent\textbf{Hierarchical features.}
In SynQT, we extract three different types of features from KSMs for classification: final output of KSM $H_i$, attention features $F^{attn}_i$, and FFN features $F^{ffn}_i$, where $i \in \{1,2,...,l\}$ and $l$ is the number of layers. All these hierarchical features contain rich information and contribute to the performance boost of SynQT, as demonstrated in Table \ref{tab:feat}. In particular, $H_i$ plays a more important role than other types of features.

\noindent\textbf{Projection in the classification head.}
We vary the projection architecture in the classification head and investigate its performance in Table \ref{tab:projection}. 
A projection of the input is necessary for the classification head, as hierarchical features across various layers may be misaligned in the representation space. However, to our surprise, a shared bottleneck structure is enough to handle these hierarchical features. The reason may be that independent projection for distinct 
features leads to overfitting, subsequently causing a slight drop in performance.

\noindent\textbf{Conditional feature aggregation.}
In the classification head, we aggregate features using the weights conditioned on  $H_l$.
We further conduct experiments over variants of SynQT with different aggregation methods including
a simple average and weighted-sum with learnable but fixed weights.
The results in Table \ref{tab:aggregation} show that conditional aggregation improves the performance of SynQT 
especially for the Structured group.

\noindent\textbf{DropFeat in the classification head.} 
We study the effect of DropFeat, which is proposed to prevent overfitting by randomly dropping features in the classification head.
As can be observed from Table \ref{tab:dropfeat}, SynQT performs worse without DropFeat, indicating that DropFeat enhances the model's generalization
\begin{wrapfigure}[16]{tr}{0.5\columnwidth}
    \centering
    \includegraphics[width=\linewidth]{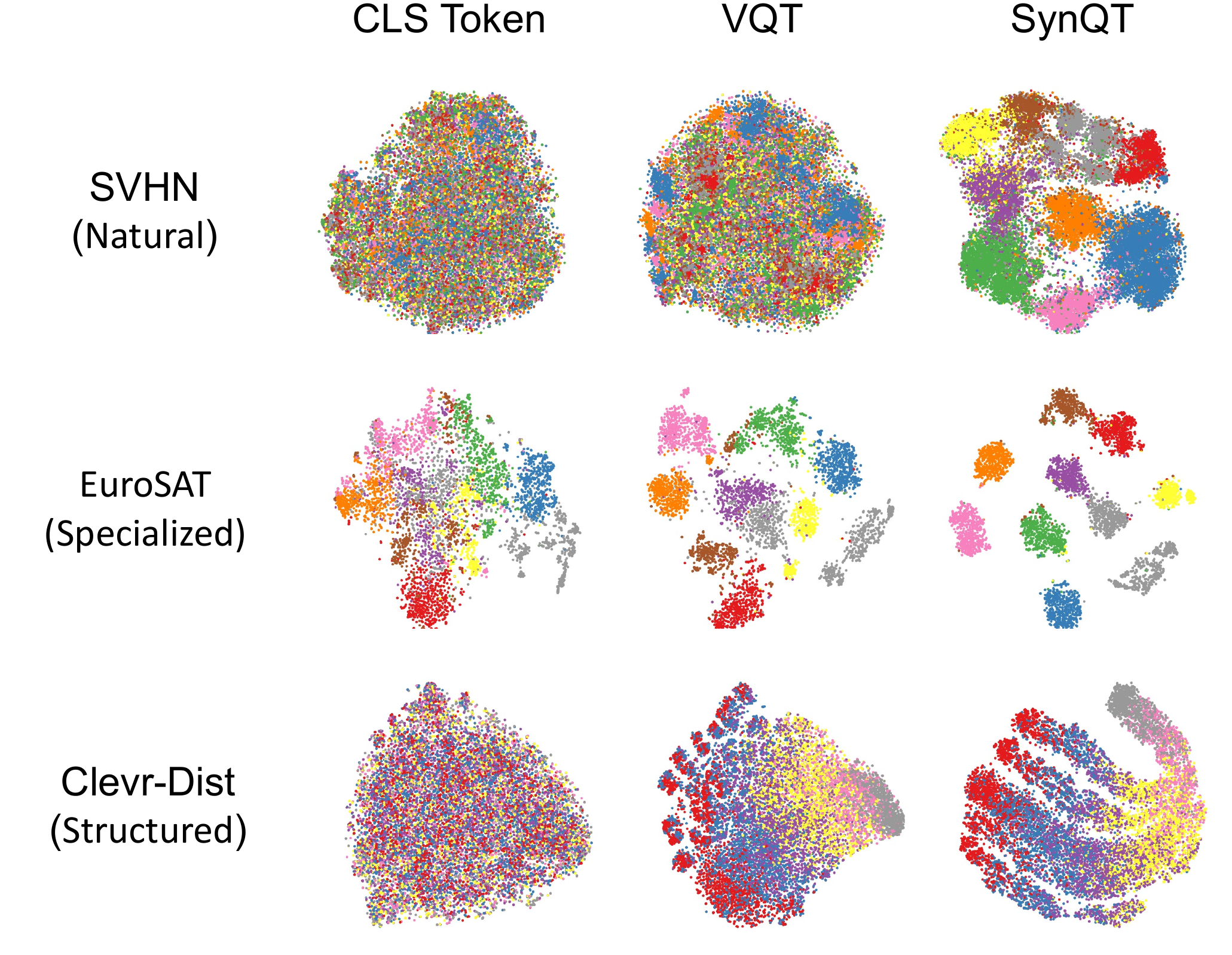}
    \caption{\textbf{t-SNE visualization on SVHN, EuroSAT, and Clevr-Dist, one from each category of VTAB-1K.} The features of our SynQT are more separable.}
    \label{fig:tsne}
\end{wrapfigure}
 in handling rich hierarchical feature input. DropFeat ensures that the model does not overly rely on a specific feature during training and encourages the model to focus on a general way to select beneficial information from these features
for downstream tasks.



\subsection{t-SNE Visualization}

We present t-SNE visualization \cite{van2008visualizing} of the CLS Token in pre-trained ViT, VQT, and our SynQT on EuroSAT, SVHN, and Clevr-Dist in Figure \ref{fig:tsne}. The visualization shows that 
our SynQT achieves better feature clustering results compared to the original CLS Token and VQT. We provide more visualization in Appendix.  
    



\section{Conclusion, Limitation, and Future Work}

In this work, we present an insightful analysis of the high memory cost of existing PETL methods from a unique perspective, $i.e.$, they entangle task-specific learning and pre-trained knowledge
utilization together. Inspired by the idea of disentanglement, we propose a new paradigm, namely SynQT, consisting of two key structures: Query Synthesis Module (QSM) and  Knowledge Extraction Module (KEM). QSM synthesizes task-specific query and then KEM extracts pre-trained knowledge in a query-only manner. Finally, a classification head is proposed to handle rich hierarchical features from KEMs and make predictions. 
Extensive experiments on VTAB-1K and FGVC demonstrate the effectiveness of SynQT under memory constraints and in an extremely low-data regime.

Despite the promising performance of our SynQT, it also has some limitations. One limitation is that SynQT requires slightly more trainable parameters than previous PETL methods due to a few projection layers and FFN operations, which is analyzed in Appendix. However, we would like to emphasize that as shown in their original papers, for VPT-Deep \cite{jia2022visual} and VQT \cite{tu2023visual}, 
straightforwardly increasing the number of tokens to use more trainable parameters can not consistently result in better performance.

We believe that our work will inspire the community to explore memory-efficient PETL along with our disentangled perspective.
Besides, to overcome the aforementioned limitation, one future direction is to develop PETL methods that achieve high accuracy and meanwhile have extreme memory efficiency and parameter efficiency.


\section*{Acknowledgments}
This work is supported in part by the National Natural Science Foundation of China under Grant 62171248, Shenzhen Science and Technology Program (JCYJ20220818101012025), and the PCNL KEY project (PCL2023AS6-1). 
%
%
\bibliographystyle{splncs04}
\bibliography{main}

\begin{thebibliography}{10}
\providecommand{\url}[1]{\texttt{#1}}
\providecommand{\urlprefix}{URL }
\providecommand{\doi}[1]{https://doi.org/#1}

\bibitem{argyriou2006multi}
Argyriou, A., Evgeniou, T., Pontil, M.: Multi-task feature learning. Advances in neural information processing systems  \textbf{19} (2006)

\bibitem{ba2016layer}
Ba, J.L., Kiros, J.R., Hinton, G.E.: Layer normalization. arXiv preprint arXiv:1607.06450  (2016)

\bibitem{bai2024badclip}
Bai, J., Gao, K., Min, S., Xia, S.T., Li, Z., Liu, W.: Badclip: Trigger-aware prompt learning for backdoor attacks on clip. In: Proceedings of the IEEE/CVF Conference on Computer Vision and Pattern Recognition. pp. 24239--24250 (2024)

\bibitem{bai2022improving}
Bai, J., Yuan, L., Xia, S.T., Yan, S., Li, Z., Liu, W.: Improving vision transformers by revisiting high-frequency components. In: European Conference on Computer Vision. pp. 1--18. Springer (2022)

\bibitem{bossard2014food}
Bossard, L., Guillaumin, M., Van~Gool, L.: Food-101--mining discriminative components with random forests. In: Computer Vision--ECCV 2014: 13th European Conference, Zurich, Switzerland, September 6-12, 2014, Proceedings, Part VI 13. pp. 446--461. Springer (2014)

\bibitem{brown2020language}
Brown, T., Mann, B., Ryder, N., Subbiah, M., Kaplan, J.D., Dhariwal, P., Neelakantan, A., Shyam, P., Sastry, G., Askell, A., et~al.: Language models are few-shot learners. NeurIPS  (2020)

\bibitem{caron2021emerging}
Caron, M., Touvron, H., Misra, I., J{\'e}gou, H., Mairal, J., Bojanowski, P., Joulin, A.: Emerging properties in self-supervised vision transformers. In: Proceedings of the IEEE/CVF international conference on computer vision. pp. 9650--9660 (2021)

\bibitem{chen2022adaptformer}
Chen, S., Ge, C., Tong, Z., Wang, J., Song, Y., Wang, J., Luo, P.: Adaptformer: Adapting vision transformers for scalable visual recognition. Advances in Neural Information Processing Systems  \textbf{35},  16664--16678 (2022)

\bibitem{deng2009imagenet}
Deng, J., Dong, W., Socher, R., Li, L.J., Li, K., Fei-Fei, L.: Imagenet: A large-scale hierarchical image database. In: 2009 IEEE conference on computer vision and pattern recognition. pp. 248--255. Ieee (2009)

\bibitem{dosovitskiy2020image}
Dosovitskiy, A., Beyer, L., Kolesnikov, A., Weissenborn, D., Zhai, X., Unterthiner, T., Dehghani, M., Minderer, M., Heigold, G., Gelly, S., et~al.: An image is worth 16x16 words: Transformers for image recognition at scale. arXiv preprint arXiv:2010.11929  (2020)

\bibitem{evci2022head2toe}
Evci, U., Dumoulin, V., Larochelle, H., Mozer, M.C.: Head2toe: Utilizing intermediate representations for better transfer learning. In: International Conference on Machine Learning. pp. 6009--6033. PMLR (2022)

\bibitem{guo2023adaptir}
Guo, H., Dai, T., Bai, Y., Chen, B., Xia, S.T., Zhu, Z.: Adaptir: Parameter efficient multi-task adaptation for pre-trained image restoration models. arXiv preprint arXiv:2312.08881  (2023)

\bibitem{guo2024mambair}
Guo, H., Li, J., Dai, T., Ouyang, Z., Ren, X., Xia, S.T.: Mambair: A simple baseline for image restoration with state-space model. arXiv preprint arXiv:2402.15648  (2024)

\bibitem{he2022masked}
He, K., Chen, X., Xie, S., Li, Y., Doll{\'a}r, P., Girshick, R.: Masked autoencoders are scalable vision learners. In: Proceedings of the IEEE/CVF conference on computer vision and pattern recognition. pp. 16000--16009 (2022)

\bibitem{houlsby2019parameter}
Houlsby, N., Giurgiu, A., Jastrzebski, S., Morrone, B., De~Laroussilhe, Q., Gesmundo, A., Attariyan, M., Gelly, S.: Parameter-efficient transfer learning for nlp. In: International Conference on Machine Learning. pp. 2790--2799. PMLR (2019)

\bibitem{hu2021lora}
Hu, E.J., Wallis, P., Allen-Zhu, Z., Li, Y., Wang, S., Wang, L., Chen, W., et~al.: Lora: Low-rank adaptation of large language models. In: International Conference on Learning Representations (2021)

\bibitem{jia2022visual}
Jia, M., Tang, L., Chen, B.C., Cardie, C., Belongie, S., Hariharan, B., Lim, S.N.: Visual prompt tuning. In: European Conference on Computer Vision. pp. 709--727. Springer (2022)

\bibitem{jie2023fact}
Jie, S., Deng, Z.H.: Fact: Factor-tuning for lightweight adaptation on vision transformer. In: Proceedings of the AAAI Conference on Artificial Intelligence. vol.~37, pp. 1060--1068 (2023)

\bibitem{jie2023revisiting}
Jie, S., Wang, H., Deng, Z.H.: Revisiting the parameter efficiency of adapters from the perspective of precision redundancy. In: Proceedings of the IEEE/CVF International Conference on Computer Vision. pp. 17217--17226 (2023)

\bibitem{kirillov2023segment}
Kirillov, A., Mintun, E., Ravi, N., Mao, H., Rolland, C., Gustafson, L., Xiao, T., Whitehead, S., Berg, A.C., Lo, W.Y., et~al.: Segment anything. arXiv preprint arXiv:2304.02643  (2023)

\bibitem{kornblith2019better}
Kornblith, S., Shlens, J., Le, Q.V.: Do better imagenet models transfer better? In: Proceedings of the IEEE/CVF conference on computer vision and pattern recognition. pp. 2661--2671 (2019)

\bibitem{krause20133d}
Krause, J., Stark, M., Deng, J., Fei-Fei, L.: 3d object representations for fine-grained categorization. In: Proceedings of the IEEE international conference on computer vision workshops. pp. 554--561 (2013)

\bibitem{larsson2016fractalnet}
Larsson, G., Maire, M., Shakhnarovich, G.: Fractalnet: Ultra-deep neural networks without residuals. In: ICLR (2017)

\bibitem{lester2021power}
Lester, B., Al-Rfou, R., Constant, N.: The power of scale for parameter-efficient prompt tuning. arXiv preprint arXiv:2104.08691  (2021)

\bibitem{li2021prefix}
Li, X.L., Liang, P.: Prefix-tuning: Optimizing continuous prompts for generation. arXiv preprint arXiv:2101.00190  (2021)

\bibitem{li2024graphadapter}
Li, X., Lian, D., Lu, Z., Bai, J., Chen, Z., Wang, X.: Graphadapter: Tuning vision-language models with dual knowledge graph. Advances in Neural Information Processing Systems  \textbf{36} (2024)

\bibitem{lian2022scaling}
Lian, D., Zhou, D., Feng, J., Wang, X.: Scaling \& shifting your features: A new baseline for efficient model tuning. Advances in Neural Information Processing Systems  \textbf{35},  109--123 (2022)

\bibitem{liu2022few}
Liu, H., Tam, D., Muqeeth, M., Mohta, J., Huang, T., Bansal, M., Raffel, C.A.: Few-shot parameter-efficient fine-tuning is better and cheaper than in-context learning. NeurIPS  (2022)

\bibitem{liu2023pre}
Liu, P., Yuan, W., Fu, J., Jiang, Z., Hayashi, H., Neubig, G.: Pre-train, prompt, and predict: A systematic survey of prompting methods in natural language processing. ACM Computing Surveys  \textbf{55}(9),  1--35 (2023)

\bibitem{liu2021p}
Liu, X., Ji, K., Fu, Y., Tam, W.L., Du, Z., Yang, Z., Tang, J.: P-tuning v2: Prompt tuning can be comparable to fine-tuning universally across scales and tasks. arXiv preprint arXiv:2110.07602  (2021)

\bibitem{liu2021swin}
Liu, Z., Lin, Y., Cao, Y., Hu, H., Wei, Y., Zhang, Z., Lin, S., Guo, B.: Swin transformer: Hierarchical vision transformer using shifted windows. In: Proceedings of the IEEE/CVF international conference on computer vision. pp. 10012--10022 (2021)

\bibitem{liu2022convnet}
Liu, Z., Mao, H., Wu, C.Y., Feichtenhofer, C., Darrell, T., Xie, S.: A convnet for the 2020s. In: Proceedings of the IEEE/CVF conference on computer vision and pattern recognition. pp. 11976--11986 (2022)

\bibitem{loshchilov2017decoupled}
Loshchilov, I., Hutter, F.: Decoupled weight decay regularization. arXiv preprint arXiv:1711.05101  (2017)

\bibitem{lu2024beyond}
Lu, Z., Bai, J., Li, X., Xiao, Z., Wang, X.: Beyond sole strength: Customized ensembles for generalized vision-language models. In: Forty-first International Conference on Machine Learning (2024)

\bibitem{van2008visualizing}
Van~der Maaten, L., Hinton, G.: Visualizing data using t-sne. Journal of machine learning research  \textbf{9}(11) (2008)

\bibitem{maji2013fine}
Maji, S., Rahtu, E., Kannala, J., Blaschko, M., Vedaldi, A.: Fine-grained visual classification of aircraft. arXiv preprint arXiv:1306.5151  (2013)

\bibitem{nilsback2006visual}
Nilsback, M.E., Zisserman, A.: A visual vocabulary for flower classification. In: 2006 IEEE computer society conference on computer vision and pattern recognition (CVPR'06). vol.~2, pp. 1447--1454. IEEE (2006)

\bibitem{parkhi2012cats}
Parkhi, O.M., Vedaldi, A., Zisserman, A., Jawahar, C.: Cats and dogs. In: 2012 IEEE conference on computer vision and pattern recognition. pp. 3498--3505. IEEE (2012)

\bibitem{pfeiffer2020adapterfusion}
Pfeiffer, J., Kamath, A., R{\"u}ckl{\'e}, A., Cho, K., Gurevych, I.: Adapterfusion: Non-destructive task composition for transfer learning. arXiv preprint arXiv:2005.00247  (2020)

\bibitem{rebuffi2017learning}
Rebuffi, S.A., Bilen, H., Vedaldi, A.: Learning multiple visual domains with residual adapters. Advances in neural information processing systems  \textbf{30} (2017)

\bibitem{shao2014transfer}
Shao, L., Zhu, F., Li, X.: Transfer learning for visual categorization: A survey. IEEE transactions on neural networks and learning systems  \textbf{26}(5),  1019--1034 (2014)

\bibitem{srivastava2014dropout}
Srivastava, N., Hinton, G., Krizhevsky, A., Sutskever, I., Salakhutdinov, R.: Dropout: a simple way to prevent neural networks from overfitting. The journal of machine learning research  \textbf{15}(1),  1929--1958 (2014)

\bibitem{sung2022lst}
Sung, Y.L., Cho, J., Bansal, M.: Lst: Ladder side-tuning for parameter and memory efficient transfer learning. Advances in Neural Information Processing Systems  \textbf{35},  12991--13005 (2022)

\bibitem{sung2022vl}
Sung, Y.L., Cho, J., Bansal, M.: Vl-adapter: Parameter-efficient transfer learning for vision-and-language tasks. In: Proceedings of the IEEE/CVF Conference on Computer Vision and Pattern Recognition. pp. 5227--5237 (2022)

\bibitem{tu2023visual}
Tu, C.H., Mai, Z., Chao, W.L.: Visual query tuning: Towards effective usage of intermediate representations for parameter and memory efficient transfer learning. In: Proceedings of the IEEE/CVF Conference on Computer Vision and Pattern Recognition. pp. 7725--7735 (2023)

\bibitem{vaswani2017attention}
Vaswani, A., Shazeer, N., Parmar, N., Uszkoreit, J., Jones, L., Gomez, A.N., Kaiser, {\L}., Polosukhin, I.: Attention is all you need. Advances in neural information processing systems  \textbf{30} (2017)

\bibitem{wan2013regularization}
Wan, L., Zeiler, M., Zhang, S., Le~Cun, Y., Fergus, R.: Regularization of neural networks using dropconnect. In: International conference on machine learning. pp. 1058--1066. PMLR (2013)

\bibitem{wang2018deep}
Wang, M., Deng, W.: Deep visual domain adaptation: A survey. Neurocomputing  \textbf{312},  135--153 (2018)

\bibitem{wang2022image}
Wang, W., Bao, H., Dong, L., Bjorck, J., Peng, Z., Liu, Q., Aggarwal, K., Mohammed, O.K., Singhal, S., Som, S., et~al.: Image as a foreign language: Beit pretraining for all vision and vision-language tasks. arXiv preprint arXiv:2208.10442  (2022)

\bibitem{yang2024not}
Yang, S., Bai, J., Gao, K., Yang, Y., Li, Y., Xia, S.T.: Not all prompts are secure: A switchable backdoor attack against pre-trained vision transfomers. In: Proceedings of the IEEE/CVF Conference on Computer Vision and Pattern Recognition. pp. 24431--24441 (2024)

\bibitem{yuan2006model}
Yuan, M., Lin, Y.: Model selection and estimation in regression with grouped variables. Journal of the Royal Statistical Society Series B: Statistical Methodology  \textbf{68}(1),  49--67 (2006)

\bibitem{zaken2021bitfit}
Zaken, E.B., Ravfogel, S., Goldberg, Y.: Bitfit: Simple parameter-efficient fine-tuning for transformer-based masked language-models. arXiv preprint arXiv:2106.10199  (2021)

\bibitem{zang2022unified}
Zang, Y., Li, W., Zhou, K., Huang, C., Loy, C.C.: Unified vision and language prompt learning. arXiv preprint arXiv:2210.07225  (2022)

\bibitem{zha2023instance}
Zha, Y., Wang, J., Dai, T., Chen, B., Wang, Z., Xia, S.T.: Instance-aware dynamic prompt tuning for pre-trained point cloud models. In: Proceedings of the IEEE/CVF International Conference on Computer Vision. pp. 14161--14170 (2023)

\bibitem{zhai2022scaling}
Zhai, X., Kolesnikov, A., Houlsby, N., Beyer, L.: Scaling vision transformers. In: Proceedings of the IEEE/CVF Conference on Computer Vision and Pattern Recognition. pp. 12104--12113 (2022)

\bibitem{zhai2019visual}
Zhai, X., Puigcerver, J., Kolesnikov, A., Ruyssen, P., Riquelme, C., Lucic, M., Djolonga, J., Pinto, A.S., Neumann, M., Dosovitskiy, A., et~al.: The visual task adaptation benchmark  (2019)

\bibitem{zhang2022NOAH}
Zhang, Y., Zhou, K., Liu, Z.: Neural prompt search. arXiv  (2022)

\bibitem{zhou2022conditional}
Zhou, K., Yang, J., Loy, C.C., Liu, Z.: Conditional prompt learning for vision-language models. In: Proceedings of the IEEE/CVF Conference on Computer Vision and Pattern Recognition. pp. 16816--16825 (2022)

\bibitem{zhou2022learning}
Zhou, K., Yang, J., Loy, C.C., Liu, Z.: Learning to prompt for vision-language models. International Journal of Computer Vision  \textbf{130}(9),  2337--2348 (2022)

\bibitem{zhu2024vision}
Zhu, L., Liao, B., Zhang, Q., Wang, X., Liu, W., Wang, X.: Vision mamba: Efficient visual representation learning with bidirectional state space model. arXiv preprint arXiv:2401.09417  (2024)

\end{thebibliography}

\appendix

\begin{center}
    \begin{Large}
        \textbf{Appendix}
    \end{Large}
\end{center}

\section{Analysis on Memory Cost and Entanglement}
\begin{wrapfigure}[13]{r}{0.4\columnwidth}
\label{tab:m_scale}
\includegraphics[width=0.35\textwidth]{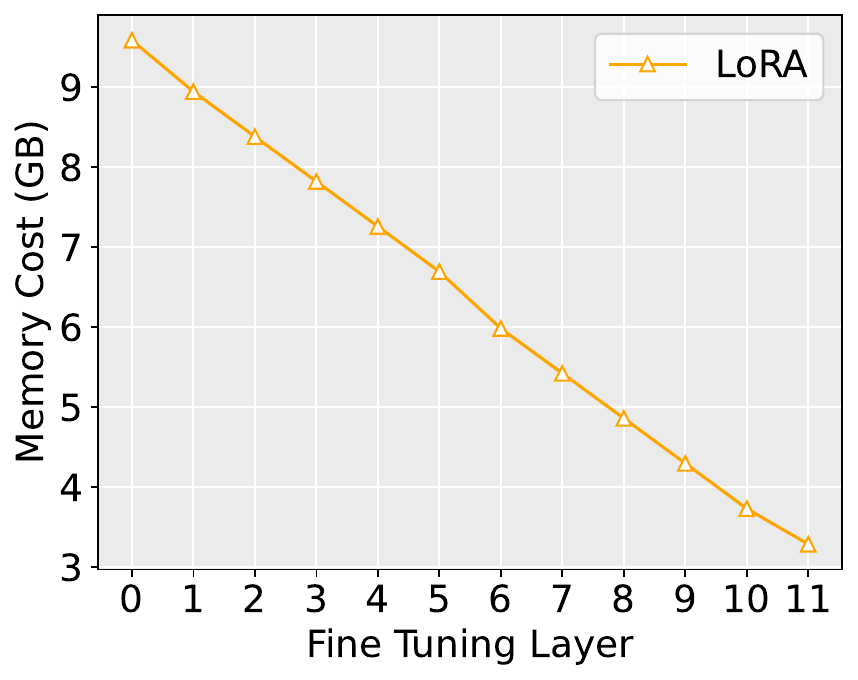}
\captionof{figure}{Memory usage with different fine-tuning layer.}
\label{fig:lora_memory}
\end{wrapfigure}
In this section, we conduct an additional experiment to investigate the connection between memory cost and entanglement.
To achieve varying degrees of entanglement, we fine-tune \textbf{only one layer} of the pre-trained model using LoRA and provide memory usage at \textbf{different positions} in Fig. \ref{fig:lora_memory}. 

When tuning the highest layer, the degree of entanglement is minimal and the features of the lower layers remain unchanged, not engaging in gradient descent, resulting in a small memory cost. As the layer being fine-tuned becomes lower, the degree of entanglement increases and the intermediate features that participate in gradient descent become heavier, leading to a larger memory cost.

\section{Analysis on Inference Cost}
The inference cost of the model is important in some real-world  applications. Compared to the original pre-trained model, SynQT, with the lightweight architectures QSM and KEM, could potentially introduce additional inference costs. Therefore, we evaluate the inference time, memory usage, and FLOPS in comparison to traditional PETL methods, and present the results along with the optimal number of tokens in Table \ref{tab:time}.

Notably, the low inference speed and high FLOPS of VPT-Shallow \cite{jia2022visual} are attributed to its optimal number of tokens on CIFAR100 reaching up to 100. In contrast, the inference speeds and FLOPS of BitFit \cite{zaken2021bitfit}, VPT-Deep \cite{jia2022visual}, VQT \cite{tu2023visual}, and SynQT are similar without any significant differences. It suggests that the important factor affecting inference speed is the number of prompt tokens rather than the additional lightweight architectures. Given that we use an extremely small number of tokens for the lightweight design of QSM, which is chosen from $\{1,4,16\}$, the overall inference speed of SynQT is comparable to VQT and the additional inference cost is affordable.

\begin{table}[htbp]
    \caption{\textbf{Inference speed analysis on CIFAR100.} We evaluate different methods with their optimal number of tokens on a single NVIDIA V100 32GB GPU and report the frames per second (fps).}
    \label{tab:time}
    \setlength{\tabcolsep}{4pt}
    \centering
    \begin{adjustbox}{width=0.9\textwidth,center}
    \begin{tabular}{c|c|c|c|c}
        \toprule
        Methods                          &  Optimal \#Tokens & Inference Speed (fps) & Memory (G) & FLOPS (G)\\
        \midrule
        BitFit \cite{zaken2021bitfit}    & -                     & 249.94 & 2.8 &     16.9                    \\
        VPT-Shallow \cite{jia2022visual} & 100                   & 172.69    & 2.9 &25.3                       \\
        VPT-Deep \cite{jia2022visual}    & 10                    & 245.76             & 2.9 & 17.7              \\
        VQT    \cite{tu2023visual}       & 1                     & 246.95             & 2.8 & 17.2              \\
        \midrule
        \baseline{\textbf{SynQT (Ours)}} & \baseline{4}          & \baseline{244.91}  & \baseline{2.9} & \baseline{17.2}              \\
        \bottomrule
    \end{tabular}
\end{adjustbox}
    \vskip-10pt
\end{table}

\section{Analysis on Feature Importance }
\begin{figure*}[t]
    \centering
    \small
    \subcaptionbox{Eurosat}{
        \includegraphics[width=0.31\textwidth]{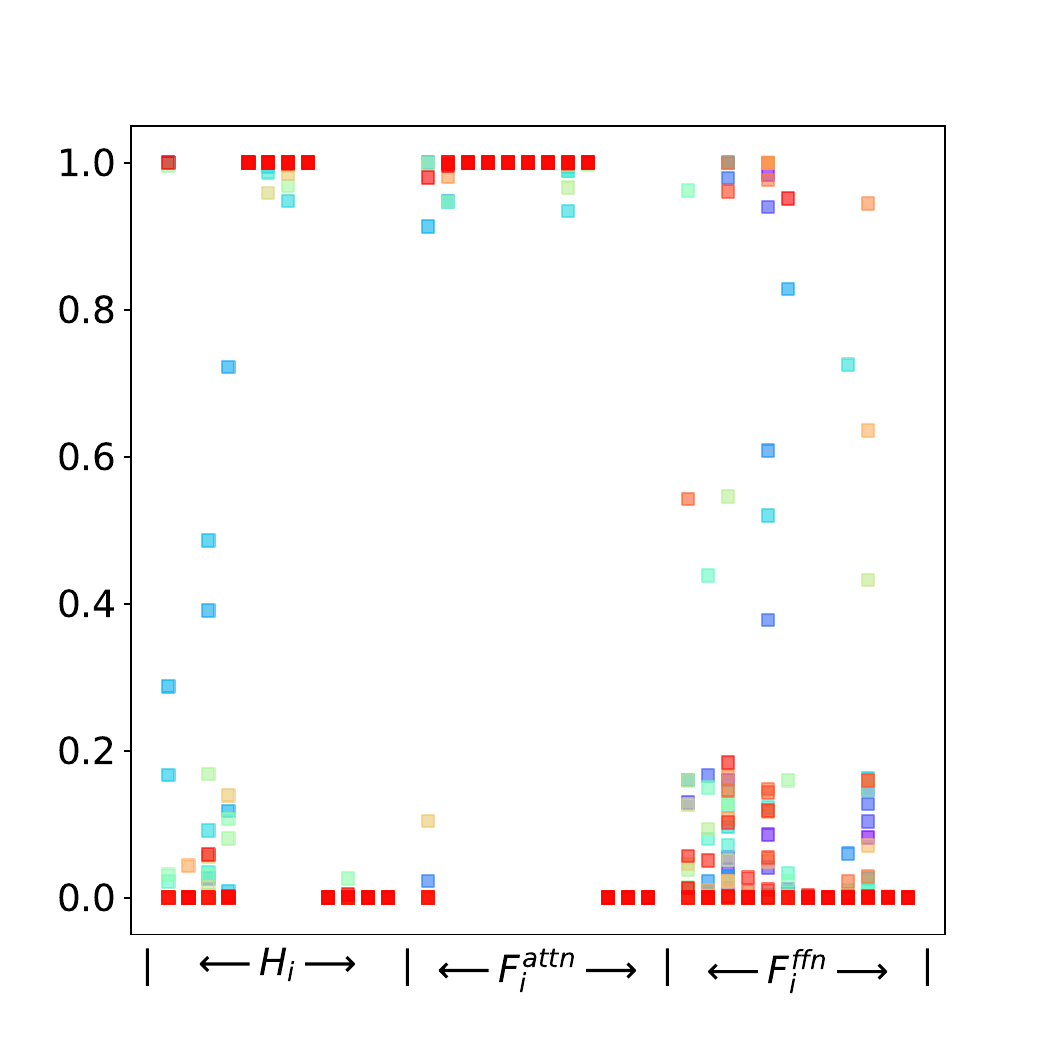}
    }
    \subcaptionbox{SVHN}{
        \includegraphics[width=0.31\textwidth]{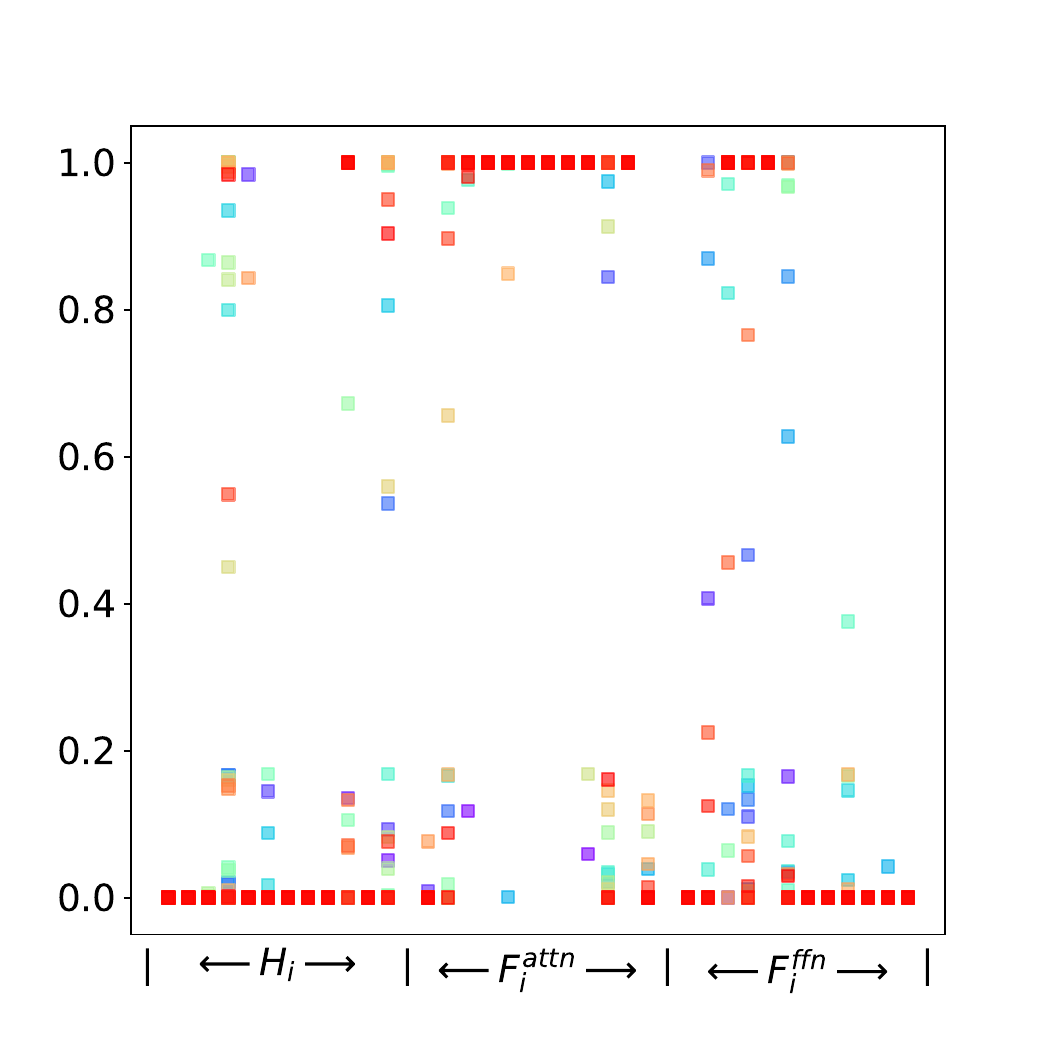}
    }
    \subcaptionbox{DMLab}{
        \includegraphics[width=0.31\textwidth]{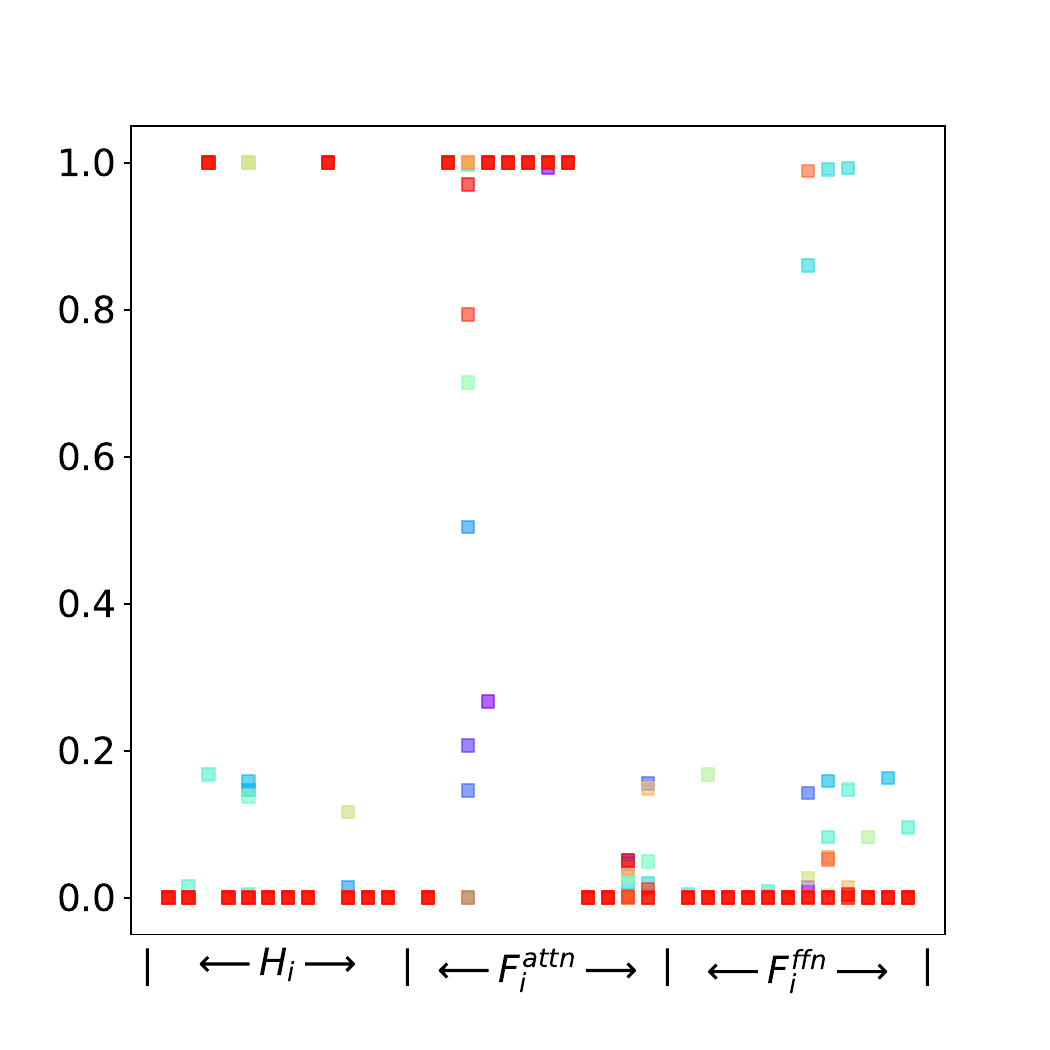}
    }
    \\

    \caption{\textbf{Feature importance visualization of 50 samples on different datasets.} The feature weights of different samples are marked with different colors. For a given sample, SynQT tends to assign relatively extreme values within the 0$\sim$1 range to distinguish useful and un-useful features.
    }
    \label{fig:weight}
\end{figure*}

The classification head in SynQT is instance-aware, by adopting the feature weights conditioned on the output of the last block.
To gain deeper insight into the feature selection within the classification head, we investigate the generated feature weights for different input samples.
Specifically, we randomly select 50 samples and visualize the importance across 36 distinct features in Figure \ref{fig:weight}. The features are divided into 3 groups:  the output of KEM $H_i$, the attention features $F_i^{attn}$ , and the FFN features $F_i^{ffn}$, where $i \in \{1,2,...,12\}$. The features within the groups are sorted by the layer index and their importance is determined by their absolute values of the feature weight.
The experiments are conducted on EuroSAT, SVHN, and DMLab.

As illustrated in Figure \ref{fig:weight}, the feature selection in the classification head assigns varying weights to different samples, thereby enabling SynQT to be instance-aware and enhancing the model's performance. 
An interesting observation is that the importance of most features is either greater than 0.8 or less than 0.2, suggesting that the classification head distinguishes between useful and non-useful features by assigning relatively extreme values within the 0$\sim$1 range for a given sample. Specifically, the number of features with a weight greater than 0.8 is noticeably fewer than those assigned a weight less than 0.2, thereby demonstrating the effectiveness of the feature selection in reducing information redundancy.


\section{Discussion on Trainable Parameters}
SynQT outperforms previous PETL methods while requiring slightly more trainable parameters due to a few projection layers and FFN operations. To validate that the superior performance of our method does not come from the additional parameters, we match the numbers of trainable parameters used in VPT-Deep \cite{jia2022visual} and VQT \cite{tu2023visual} with ours, by increasing their number of tokens to 300.
We term the VPT-Deep and VQT with more trainable parameters as VPT-Deep$^\dagger$ and VQT$^\dagger$, respectively.
The comparison results are provided in Table \ref{tab:parameters}.

\begin{table*}[t]
    \vspace{-0.1in}
    \centering
    \caption{\textbf{Comparison with versions of VPT-Deep and VQT that employ a larger number of tokens (VPT-Deep$^\dagger$ and VQT$^\dagger$) on the VTAB-1K benchmark, with ViT-B/16 pre-trained on ImageNet-21K.} ``\# Param'' specifies the number of trainable parameters in backbones. Average accuracy are averaged over group-wise mean values.}
    \label{tab:parameters}
    \setlength{\tabcolsep}{3pt}
    \resizebox{\textwidth}{!}{
        \begin{tabular}{p{2.4cm}<{}p{1cm}<{\centering}|
            p{0.75cm}<{\centering}p{0.75cm}<{\centering}p{0.75cm}<{\centering}p{0.75cm}<{\centering}p{0.75cm}<{\centering}p{0.75cm}<{\centering}p{0.75cm}<{\centering}|p{0.75cm}<{\centering}p{0.75cm}<{\centering}p{0.75cm}<{\centering}p{0.75cm}<{\centering}|p{0.75cm}<{\centering}p{0.75cm}<{\centering}p{0.75cm}<{\centering}p{0.75cm}<{\centering}p{0.75cm}<{\centering}p{0.75cm}<{\centering}p{0.75cm}<{\centering}p{0.75cm}<{\centering}|p{0.75cm}<{\centering}}
            \toprule[1.5pt]
            \multicolumn{2}{c|}{}                     & \multicolumn{7}{c|}{\textbf{Natural}}                         & \multicolumn{4}{c|}{\textbf{Specialized}} & \multicolumn{8}{c|}{\textbf{Structured}}                                                                                                                                                                                                                                                                                                 \\
                                                      & \multicolumn{1}{c|}{{\rotatebox[origin=c]{90}{\# Param (M)}}}
                                                      & \multicolumn{1}{c}{{\rotatebox[origin=c]{90}{CIFAR100}}}
                                                      & \multicolumn{1}{c}{{\rotatebox[origin=c]{90}{Caltech101}}}
                                                      & \multicolumn{1}{c}{{\rotatebox[origin=c]{90}{DTD}}}
                                                      & \multicolumn{1}{c}{{\rotatebox[origin=c]{90}{Flower102}}}
                                                      & \multicolumn{1}{c}{{\rotatebox[origin=c]{90}{Pets}}}
                                                      & \multicolumn{1}{c}{{\rotatebox[origin=c]{90}{SVHN}}}
                                                      & \multicolumn{1}{c|}{{\rotatebox[origin=c]{90}{Sun397}}}
                                                      & \multicolumn{1}{c}{{\rotatebox[origin=c]{90}{Camelyon}}}
                                                      & \multicolumn{1}{c}{{\rotatebox[origin=c]{90}{EuroSAT}}}
                                                      & \multicolumn{1}{c}{{\rotatebox[origin=c]{90}{Resisc45}}}
                                                      & \multicolumn{1}{c|}{{\rotatebox[origin=c]{90}{Retinopathy}}}
                                                      & \multicolumn{1}{c}{{\rotatebox[origin=c]{90}{Clevr-Count}}}
                                                      & \multicolumn{1}{c}{{\rotatebox[origin=c]{90}{Clevr-Dist}}}
                                                      & \multicolumn{1}{c}{{\rotatebox[origin=c]{90}{DMLab}}}
                                                      & \multicolumn{1}{c}{{\rotatebox[origin=c]{90}{KITTI-Dist}}}
                                                      & \multicolumn{1}{c}{{\rotatebox[origin=c]{90}{dSpr-Loc}}}
                                                      & \multicolumn{1}{c}{{\rotatebox[origin=c]{90}{dSpr-Ori}}}
                                                      & \multicolumn{1}{c}{{\rotatebox[origin=c]{90}{sNORB-Azim}}}
                                                      & \multicolumn{1}{c|}{{\rotatebox[origin=c]{90}{sNORB-Ele}}}
                                                      & \multicolumn{1}{c}{{\rotatebox[origin=c]{90}{Average}}}                                                                                                                                                                                                                                                                                                                                                                                              \\
            \specialrule{0em}{1pt}{1pt}
            \hline
            \specialrule{0em}{1pt}{1pt}
            VPT-Deep  \cite{jia2022visual}            & 0.17                                                          & \textbf{78.8}                             & \textbf{90.8}                            & 65.8          & 98.0          & 88.3          & \textbf{78.1} & 49.6          & 81.8          & 96.1          & 83.4          & 68.4          & 68.5          & 60.0          & 46.5          & 72.8          & 73.6          & 47.9          & \textbf{32.9} & \textbf{37.8} & 72.0          \\
            \hline
            \specialrule{0em}{1pt}{1pt}

            VPT-Deep$^\dagger$  \cite{jia2022visual}        & 2.84                                                          & 26.6                                      & 69.5                                     & 50.9          & 82            & 63.6          & 27.4          & 18.7          & 77.6          & 92.1          & 72.1          & 74.3          & 53.7          & 39.3          & 32.9          & 70.5          & 14.0          & 11.0          & 9.9           & 25.9          & 53.2          \\

            \hline
            \specialrule{0em}{1pt}{1pt}
            VQT  \cite{tu2023visual}                  & 0.09                                                          & 66.3                                      & 89.9                                     & 67.8          & 97.9          & 84.7          & 79.9          & 45.5          & 79.0          & 95.2          & 80.9          & 74.7          & 46.7          & 61.6          & 45.1          & 63.6          & 62.9          & 32.1          & 30.0          & 28.8          & 68.3          \\
            \hline
            \specialrule{0em}{1pt}{1pt}

            VQT$^\dagger$ \cite{tu2023visual}               & 2.84                                                          & 66.5                                      & 88.3                                     & 68.1          & 98.2          & 88.4          & 58.8          & 50.4          & 78.3          & 95.5          & 79.3          & 74.2          & 53.7          & 58.1          & 39.0          & 67.5          & 30.8          & 15.1          & 14.5          & 26.9          & 64.7          \\

            \hline
            \specialrule{0em}{1pt}{1pt}

            \rowcolor{lightgray}\textbf{SynQT (Ours)} & 2.73                                                          & 70.9                                      & 89.7                                     & \textbf{68.8} & \textbf{98.5} & \textbf{89.6} & 77.8          & \textbf{50.6} & \textbf{82.3} & \textbf{96.7} & \textbf{83.5} & \textbf{75.2} & \textbf{71.8} & \textbf{62.7} & \textbf{48.5} & \textbf{75.4} & \textbf{74.1} & \textbf{49.0} & 31.7          & 36.1          & \textbf{72.9}
            \\
            \bottomrule[1.5pt]
        \end{tabular}
    }
\end{table*}

We can see that increasing the number of tokens does not consistently yield a performance improvement across different datasets but may result in a significant performance drop. The observation is consistent with that in their original papers. That is, the optimal prompt length should not be too large, as the 
increased model complexity may cause overfitting due to the limited training data on VTAB-1K.
Our results suggest that the key to the success of our SynQT is not the increased trainable parameters but the appropriate designs of our QSM and KEM.

\section{Comparison among SynQT, Head2Toe, and LST}
We further present a comparison among SynQT,  Head2Toe \cite{evci2022head2toe}, and LST \cite{sung2022lst} on the VTAB-1K benchmark, using ViT-B/16 pre-trained on ImageNet-1K, as shown in Table \ref{tab:head2toe}.  Head2Toe is another method that utilizes intermediate representations for classification and simply applies an averaging over these features to reduce dimension. The input of Head2Toe consists of features from four distinct stages: after the layer normalization, after the Multi-head Attention block, inside, and after the MLP block.
LST proposes a side network to sequentially take intermediate features as inputs for memory saving.  LST also initializes the ladder-side network based on structural pruning to enhance performance. 

Compared to Head2Toe and LST from the results, SynQT achieves a significant performance improvement.
Specifically, SynQT outperforms Head2Toe on 16 out of 19 tasks and surpasses LST by 3.3\% on average, indicating that
the intermediate representations extracted by the task-specific synthesized query are more powerful than features derived from the original pre-trained model.

\section{Discussion on Scaling Factors}
In SynQT, we introduce two scaling factors $s'$ and $s''$ to control the information flow in the QSM. A recent work \cite{lian2022scaling} indicates that shifting the intermediate features helps close the gap between pre-training and the downstream task. 
Inspired by \cite{lian2022scaling},  in our case, we make use of these scaling factors to cope with the significant gaps between pre-training and various downstream tasks, e.g., 19 datasets in VTAB-1K.

To have a better understanding of these scaling factors, we further set them as 1 and report the results in Table \ref{tab:scale}. The results indicate that the scaling operation significantly improves the performance across different datasets. In particular, SynQT shows a 5.9\% improvement in the Structured category and an average improvement of 2.4\% due to these scaling factors.

\section{Discussion on Frozen KEM}
In SynQT, we reuse the original model's weights in the KEM due to their ability to effectively extract features from the intermediate features. 
In our case, we keep them frozen to leverage their abilities and preserve the feature interaction in the original space. 
Additionally, the frozen KEM is beneficial for PETL methods, effectively reducing the number of trainable parameters during training and preventing overfitting on downstream small datasets. 

We also provide the comparison with a variant incorporating trainable KEM, whose training complexity would be similar to full-tuning. As shown in Table \ref{tab:kem}, a trainable KEM would not benefit SynQT due to the large number of trainable parameters and the limited training samples available in the downstream tasks.

\begin{table*}[t]
    \vspace{-0.1in}
    \centering
    \caption{\textbf{Comparison among SynQT, LST and Head2Toe on VTAB-1K benchmark with ViT-B/16  pre-trained on ImageNet-1K.} Average accuracy are averaged over group-wise mean values.}
    \vspace{-0.5em}
    \label{tab:head2toe}
    \setlength{\tabcolsep}{3pt}
    \resizebox{\textwidth}{!}{
        \begin{tabular}{p{2.5cm}<{}|p{0.6cm}<{\centering}
            p{0.6cm}<{\centering}p{0.6cm}<{\centering}p{0.6cm}<{\centering}p{0.6cm}<{\centering}p{0.6cm}<{\centering}p{0.6cm}<{\centering}|p{0.6cm}<{\centering}p{0.6cm}<{\centering}p{0.6cm}<{\centering}p{0.6cm}<{\centering}|p{0.6cm}<{\centering}p{0.6cm}<{\centering}p{0.6cm}<{\centering}p{0.6cm}<{\centering}p{0.6cm}<{\centering}p{0.6cm}<{\centering}p{0.6cm}<{\centering}p{0.6cm}<{\centering}|p{0.6cm}<{\centering}}
            \toprule[1.5pt]
            \multicolumn{1}{c|}{}                     & \multicolumn{7}{c|}{\textbf{Natural}}                        & \multicolumn{4}{c|}{\textbf{Specialized}} & \multicolumn{8}{c|}{\textbf{Structured}}                                                                                                                                                                                                                                                                                 \\
            \multicolumn{1}{c|}{{\rotatebox[origin=c]{90}{}}}      
                                                      & \multicolumn{1}{c}{{\rotatebox[origin=c]{90}{CIFAR100}}}
                                                      & \multicolumn{1}{c}{{\rotatebox[origin=c]{90}{Caltech101}}}
                                                      & \multicolumn{1}{c}{{\rotatebox[origin=c]{90}{DTD}}}
                                                      & \multicolumn{1}{c}{{\rotatebox[origin=c]{90}{Flower102}}}
                                                      & \multicolumn{1}{c}{{\rotatebox[origin=c]{90}{Pets}}}
                                                      & \multicolumn{1}{c}{{\rotatebox[origin=c]{90}{SVHN}}}
                                                      & \multicolumn{1}{c|}{{\rotatebox[origin=c]{90}{Sun397}}}
                                                      & \multicolumn{1}{c}{{\rotatebox[origin=c]{90}{Camelyon}}}
                                                      & \multicolumn{1}{c}{{\rotatebox[origin=c]{90}{EuroSAT}}}
                                                      & \multicolumn{1}{c}{{\rotatebox[origin=c]{90}{Resisc45}}}
                                                      & \multicolumn{1}{c|}{{\rotatebox[origin=c]{90}{Retinopathy}}}
                                                      & \multicolumn{1}{c}{{\rotatebox[origin=c]{90}{Clevr-Count}}}
                                                      & \multicolumn{1}{c}{{\rotatebox[origin=c]{90}{Clevr-Dist}}}
                                                      & \multicolumn{1}{c}{{\rotatebox[origin=c]{90}{DMLab}}}
                                                      & \multicolumn{1}{c}{{\rotatebox[origin=c]{90}{KITTI-Dist}}}
                                                      & \multicolumn{1}{c}{{\rotatebox[origin=c]{90}{dSpr-Loc}}}
                                                      & \multicolumn{1}{c}{{\rotatebox[origin=c]{90}{dSpr-Ori}}}
                                                      & \multicolumn{1}{c}{{\rotatebox[origin=c]{90}{sNORB-Azim}}}
                                                      & \multicolumn{1}{c|}{{\rotatebox[origin=c]{90}{sNORB-Ele}}}
                                                      & \multicolumn{1}{c}{{\rotatebox[origin=c]{90}{Average}}}                                                                                                                                                                                                                                                                                                                                                                             \\
            \specialrule{0em}{1pt}{1pt}
            \hline
            \specialrule{0em}{1pt}{1pt}
            Head2Toe \cite{evci2022head2toe}          & 58.2                                                         & 87.3                                      & 64.5                                     & 85.9          & 85.4          & \textbf{82.9} & 35.1          & 81.2          & 95.0          & 79.9          & 74.1          & 49.3          & 58.4          & 41.6          & 64.4          & 53.3          & 32.9          & \textbf{33.5} & \textbf{39.4} & 63.3          \\
            LST \cite{sung2022lst} & 51.8&
83.7&
62.0&
\textbf{93.2}&
78.9&
77.1&
28.7&
80.5&
\textbf{96.6}&
79.6&
75.1&
\textbf{76.3}&
61.0&
44.1&
73.4&
\textbf{73.9}&
35.6&
27.1&
35.9&
68.1
\\
            \hline
            \specialrule{0em}{1pt}{1pt}


            \rowcolor{lightgray}\textbf{SynQT (Ours)} & \textbf{59.2}                                                & \textbf{89.7}                             & \textbf{66.2}                            & 91.9 & \textbf{88.9} & 77.4          & \textbf{39.0} & \textbf{84.0} & \textbf{96.6} & 82.9          & \textbf{75.4} & {68.1} & \textbf{60.2} & \textbf{47.9} & 76.9          & 73.8 & \textbf{52.4} & {32.4}        & {37.7}        & \textbf{71.4} \\
            \bottomrule[1.5pt]
        \end{tabular}
    }
    \vspace{-0.06in}
\end{table*}

\begin{table}[t]
\begin{minipage}{0.48\textwidth}
\centering
\captionof{table}{Comparison with variant having a scale equal to 1 (SynQT$^\dagger$).}
    \label{tab:scale}
    \setlength\tabcolsep{1pt}
\resizebox{\columnwidth}{!}{
    \begin{tabular}{lcccc}
        \toprule 
        Methods & Natural & Specialized & Structured  & Average \\
        \midrule
        SynQT$^\dagger$           & 77.7            & 83.6                     & 50.3                     & 70.5                     \\
        \textbf{SynQT} & \textbf{78.0} & \textbf{84.4} & \textbf{56.2} & \textbf{72.9}         \\
        \bottomrule
    \end{tabular}
}
\end{minipage}
\hspace{1em}
\begin{minipage}{0.48\textwidth}
\centering
\captionof{table}{Comparison with variants having trainable KEM (SynQT$^\ddag$).}
    \label{tab:kem}
    \setlength\tabcolsep{1pt}
\resizebox{\columnwidth}{!}{
    \begin{tabular}{lcccc}
        \toprule
         Methods & Natural & Specialized & Structured  & Average \\
        \midrule
                SynQT$^\ddag$          & 60.5&
77.7&
36.6&
58.3  \\
        \textbf{SynQT} & \textbf{78.0} & \textbf{84.4} & \textbf{56.2} & \textbf{72.9}         \\
        \bottomrule
    \end{tabular}
}
\end{minipage}
\end{table}

\begin{figure}
    \centering
    \includegraphics[width=0.65\textwidth, page=2]{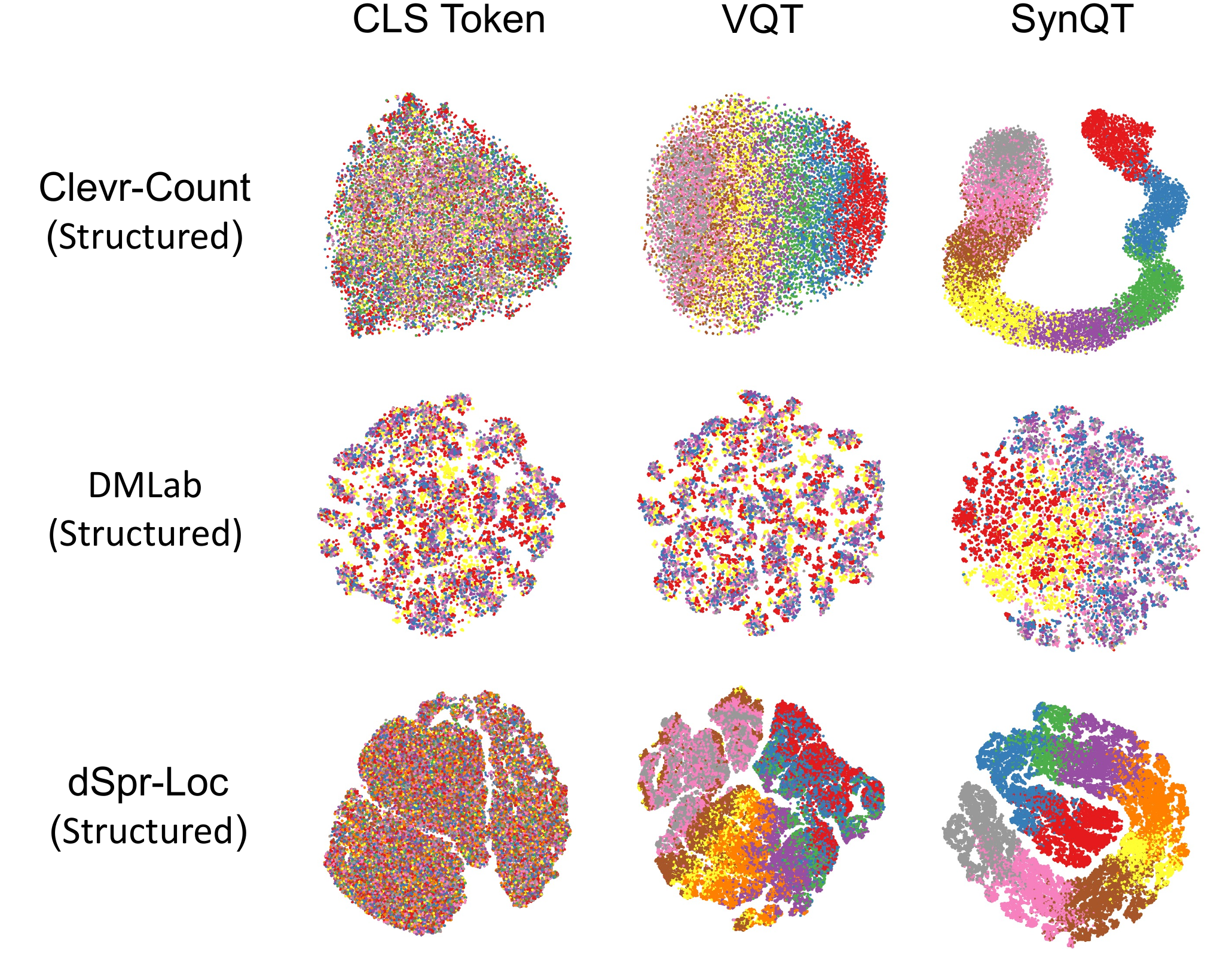}
    \caption{\textbf{t-SNE visualization on more datasets including Clevr-Count, DMLab, and dSpr-Loc.}
    }
    \label{fig:tsne_appendix}
\end{figure}

\section{t-SNE Visualization on More Datasets}
We provide t-SNE \cite{van2008visualizing} visualization on more datasets in Figure \ref{fig:tsne_appendix}. Similar to the visualization in the main manuscript, it further shows that the features obtained by SynQT are more separable compared to the original CLS Token and VQT.



\end{document}